\documentclass[10pt,twocolumn,letterpaper]{article}

\usepackage[pagenumbers]{cvpr} %
\usepackage[inkscapelatex=false]{svg}
\usepackage{overpic}
\usepackage[pdftex,outline]{contour}
\def\cov#1#2{#1{\boldsymbol{\Sigma}}_{#2}}
\def\covadj{\boldsymbol{\Sigma}_{N}}
\def\identity{\mathbb{I}}
\def\alphaT#1#2{#1{\alpha}_{#2}}
\def\gammaT#1#2{#1{\gamma}_{#2}}
\def\U{\boldsymbol{U}}
\def\Eigenvalues#1#2{#1{\boldsymbol{\Lambda}}_{#2}}
\def\mean#1#2{#1{\boldsymbol{\mu}}_{#2}}

\def\x#1#2{#1{\boldsymbol{x}}_{#2}}
\def\epsi#1#2{#1{\boldsymbol{\epsilon}}_{#2}}
\def\forward{p}
\def\posterior{q}
\def\prior{\forward}
\def\truedist{\forward}

\def\pastvar#1{\textbf{X}_{#1}}
\def\futurevar#1{\textbf{Y}_{#1}}
\def\motionvar#1{\textbf{M}_{#1}}

\def\nsamples{N}
\def\pastwindow{P} %
\def\futurewindow{F}
\def\pose#1#2{{\textbf{p}}^{#1}_{#2}}
\def\predpose#1#2{\tilde{\textbf{p}}^{#1}_{#2}}
\def\numjoints{J}

\def\encnet{\mathrm{{e}}} %
\def\decnet{\mathrm{{d}}}
\def\diffnet{\mathrm{{g}}}
\def\latentvar#1{\boldsymbol{z}_{#1}}

\def\latentdim{L}

\def\tglinear#1{f_{#1}}

\def\reachmatrix{\mathbf{R}}
\def\reachmatrixhip{\reachmatrix^{hip}}
\def\adjmatrix{\mathbf{A}}

\def\timesteps{\tau}

\def\W#1{\mathbf{W}^{#1}}
\def\feat#1{\mathbf{f}^{#1}}

\def\numlimbs{\mathrm{B}}

\def\ourmethod{SkeletonDiffusion}

\definecolor{pastgreen}{rgb}{0.0, 0.59019, 0.362745}
\definecolor{futurewhite}{rgb}{0.9431372,0.9431372, 0.9431372}

\definecolor{bluepred}{rgb}{0.0, 0.0, 0.960784}

\definecolor{futurewhite2}{rgb}{0.49,0.49, 0.49}
\definecolor{bluepred2}{rgb}{0.184,0.255, 0.533}
\usepackage{etoc}
\usepackage{multirow}

\newcommand{\methodname}[0]{SkeletonDiffusion}

\definecolor{cvprblue}{rgb}{0.21,0.49,0.74}
\usepackage[pagebackref,breaklinks,colorlinks,allcolors=cvprblue]{hyperref}
\usepackage{pst-text}

\def\paperID{} %
\def\confName{CVPR}
\def\confYear{2025}

\title{Nonisotropic Gaussian Diffusion for Realistic 3D Human Motion Prediction}

\author{
Cecilia Curreli$^{1, 2}$ \hspace{0.8cm} Dominik Muhle$^{1,2}$ \hspace{0.8cm} Abhishek Saroha$^{1,2}$ \\
Zhenzhang Ye$^{1}$\hspace{0.7cm} Riccardo Marin$^{1,2}$ \hspace{0.7cm}  Daniel Cremers$^{1,2}$ \\
\small{$^1$Technical University of Munich \hspace{1cm} $^2$Munich Center for Machine Learning}\hspace{1cm} 
}

\newcommand\YUGE{\fontsize{100}{100}\selectfont}

\begin{document}
\twocolumn[{
    \renewcommand\twocolumn[1][]{#1}%
    \maketitle
    \vspace{-0.6cm}
    \begin{center}
    \centering%
    \captionsetup{type=figure}
    
     \begin{overpic}[trim=1cm 1cm 0cm 0.5cm,clip, width=0.9\linewidth]{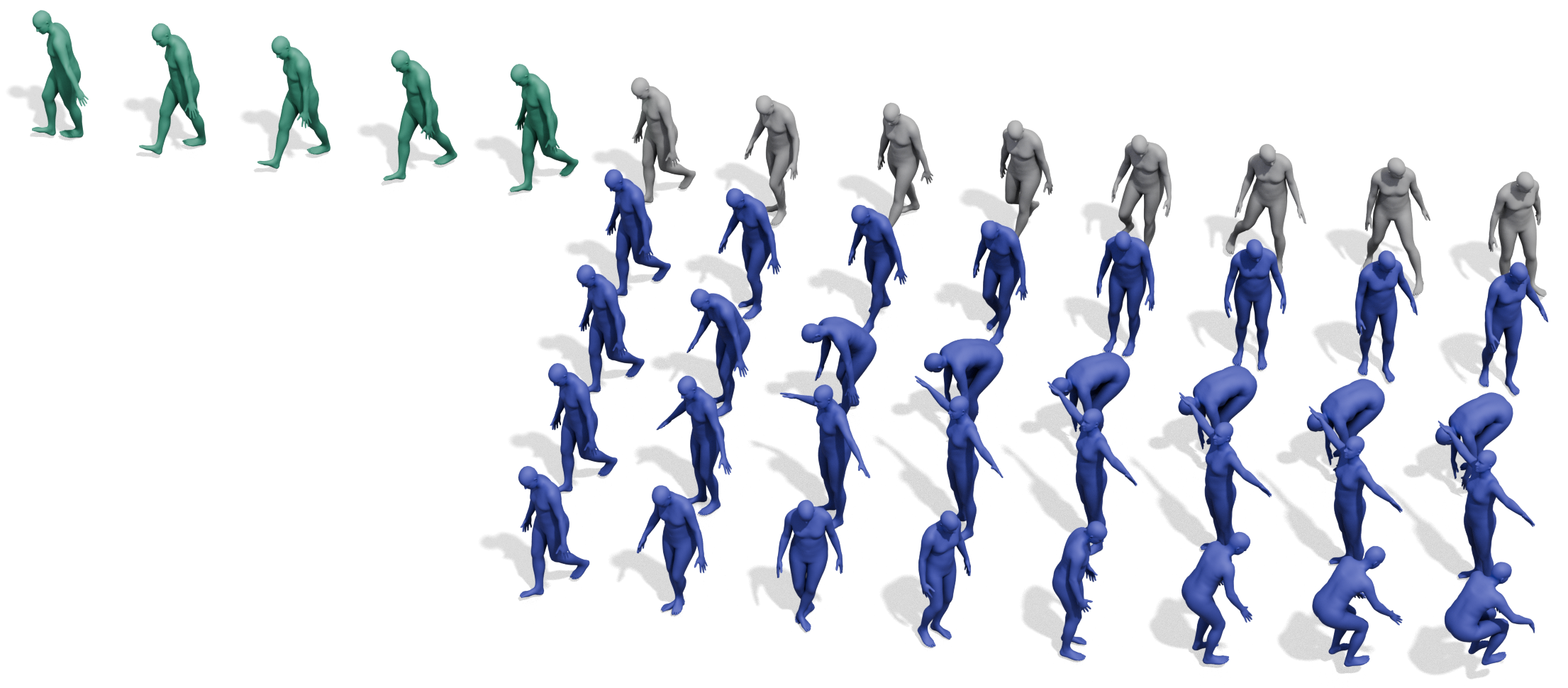} %
        \put(20,42){\LARGE \rotatebox{-6}{\color{pastgreen}\textbf{Past}}}
        \put(60,38){\LARGE \rotatebox{-6}{\color{futurewhite2}\textbf{GT Future}}}
        \put(3,24){\LARGE \rotatebox{-6}{\color{bluepred2}\textbf{Predicted}}}
        \put(4.5,20.5){\LARGE \rotatebox{-6}{\color{bluepred2}\textbf{futures}}}

        \put(12.3,21){\YUGE \rotatebox{-18}{\color{bluepred2} $\Biggl\{$}}
    
        \put(29,28){\rotatebox{-9}{\color{bluepred2} \textbf{Closest}}}
        \put(25,23.1){\rotatebox{-9}{\color{bluepred2} \textbf{Diverse \#1}}}
        \put(23,16.9){\rotatebox{-9}{\color{bluepred2} \textbf{Diverse \#2}}}
        \put(21.0,11){\rotatebox{-9}{\color{bluepred2} \textbf{Diverse \#3}}}
        
    \end{overpic}

      \captionof{figure}{\textbf{\methodname~generates \textbf{\textcolor{bluepred2}{futures}} that are simultaneously \underline{diverse} and \underline{realistic}}. With a nonisotropic diffusion formulation reflecting the skeleton structure, we predict motions that are plausible and semantically coherent with the input  \textbf{\textcolor{pastgreen}{past}} while being highly diverse. Here, we show the most diverse ensemble of three motions including the prediction closest to the \textbf{\textcolor{futurewhite2}{ground truth}} among 50 generated futures. }
    \label{fig:teaser}
    \end{center}
}]
\begin{abstract}
Probabilistic human motion prediction aims to forecast multiple possible future movements from past observations. While current approaches report high diversity and realism, they often generate motions with undetected limb stretching and jitter. To address this, we introduce \emph{\ourmethod}, a latent diffusion model that embeds an explicit inductive bias on the human body within its architecture and training. We present a new nonisotropic Gaussian diffusion formulation that aligns with the natural kinematic structure of the human skeleton and models relationships between body parts. Results show that our approach outperforms isotropic alternatives, consistently generating realistic predictions while avoiding artifacts such as limb distortion. Additionally, we identify a limitation in commonly used diversity metrics, which may favor models that produce inconsistent limb lengths within the same sequence. \emph{\ourmethod} sets a new benchmark on three real-world datasets, outperforming various baselines across multiple evaluation metrics. We release the code on our \href{https://ceveloper.github.io/publications/skeletondiffusion/}{project page}. 
\end{abstract}

\vspace{-12pt}
\section{Introduction}
\label{sec:intro}

In this work, we address the problem of predicting human motion based on observed past movements, known as Human Motion Prediction (HMP). Specifically, from a temporal sequence of human joint positions, we aim to forecast their evolution in subsequent frames.  HMP is a relevant problem for various real-world applications \cite{zhang2021we, yang2023neural, yuan2019diverse, xu2023interdiff, xu2023stochastic, bhattacharyya2018accurate, liu2023intention, troje2002decomposing, ju2023human} and the key enabler of various downstream tasks \cite{rudenko2020human, andrist2015look}: autonomous driving \cite{paden2016survey}, healthcare \cite{taylor2020intelligent}, assistive robotics \cite{lee2022robot, teramae2017emg}, human-robot interaction \cite{butepage2017anticipating, cao2020long, lee2022robot, gui2018teaching}, and virtual reality or animation creation \cite{van2010real}. %
The task can be formulated as a deterministic regression by predicting the most likely future motion~\cite{mao2020history, ma2022progressively, dang2021msr, fragkiadaki2015recurrent, jain2016structural, martinez2017human, gui2018adversarial, pavllo2018quaternet, liu2019towards, li2018convolutional, medjaouri2022hr, aksan2021spatio, cai2020learning, martinez2021pose, guo2023back}. %
However, many applications~\cite{paden2016survey,rudenko2020human, butepage2017anticipating, cao2020long, lee2022robot, gui2018teaching} require considering the inherent uncertainty of future movements. \emph{Stochastic Human Motion Prediction} (SHMP) methods aim to learn a probability distribution over possible future motions. %
Once models are capable of representing multiple futures, the challenge lies in generating \textit{diverse} yet \textit{realistic} predictions. %
In our study, we observed that often diversity in the results comes at the cost of favoring physically unfeasible movements \cite{barquero2023belfusion}, such as velocity irregularities between frames (e.g., jittering or shaking) or inconsistent joint positions (e.g., changing bone lengths between frames). %
We believe this phenomenon to be a direct consequence of the lack of a proper inductive bias on the human skeletal structure. We present \ourmethod, a latent diffusion model encoding this bias explicitly on both architecture and training. 

First, we consider the skeleton structure and joint categories throughout the entire network, and build our architecture end-to-end on top of Graph Convolutional Networks (GCNs). 
In contrast, existing SHMP approaches either ignore the skeleton’s graph structure \cite{barquero2023belfusion, zhang2021we, yuan2020dlow, chen2023humanmac} or only leverage it at intermediate stages~\cite{mao2021generating, dang2022diverse, suncomusion, wei2023human}.
Second,  we model the generative strategy to integrate the explicit bias. Similarly to the recent advances in SHMP based on diffusion models \cite{barquero2023belfusion, suncomusion, chen2023humanmac, wei2023human}, we opt for latent diffusion \cite{rombach2022highresolution}. 
However,  we replace the conventional isotropic Gaussian diffusion training \cite{ho2020denoising} with a novel \emph{nonisotropic} formulation that accounts for joint relations directly in the generation process: the HMP problem is defined by the skeleton kinematic graph, and we exploit this knowledge to define a fixed non-diagonal noise covariance for the diffusion process. To the best of our knowledge, this is the first nonisotropic diffusion process to support a non i.i.d. latent space and reflect the dependencies \emph{among} components (joints) according to the given problem structure (skeleton kinematic).
Despite demonstrating its usefulness in the skeletal domain, its applicability can be broader and touch all the domains where the conventional i.i.d. noise assumption may not hold.

We evaluate SkeletonDiffusion against the state-of-the-art on a large MoCap dataset (AMASS \cite{mahmood2019amass}), noisy data obtained by external camera tracking (FreeMan \cite{Wang2023freeman}), and in a zero-shot setting (3DPW \cite{vonMarcard2018}). We showcase consistently improved performance by generating realistic and diverse predictions (\cref{fig:teaser}) with the least amount of stretching and jittering of bone lengths (\textit{body realism}).
In summary, our contributions are:
\begin{itemize}
    \item We derive the first nonisotropic Gaussian diffusion formulation for a structural problem, which comprehends a detailed mathematical derivation and the required equations for training and inference.
    \item We propose \ourmethod, a latent diffusion model for SHMP that explicitly incorporates end-to-end the skeleton structure through the adjacency matrix in the graph architecture and the diffusion training. 
    \item We conduct extensive analyses and demonstrate \ourmethod's state-of-the-art performance on multiple datasets. Our results demonstrate issues overlooked by previous methods (e.g., limbs' stretching, jittering) and highlight the need for new realism and diversity metrics.
\end{itemize}

\section{Related Work}
\label{sec:related_works}
\subsection{Human Motion Prediction}
Probabilistic HMP has been addressed via generative adversarial networks \cite{barsoum2018hp, kundu2019bihmp, liu2021aggregated}, variational autoencoders (VAE) \cite{walker2017pose, yan2018mt, cai2021unified, mao2021generating, mao2021generating, gu2024learning, dang2022diverse, yuan2020dlow, xu2022diverse}, and more recently diffusion models \cite{suncomusion, chen2023humanmac, barquero2023belfusion, saadatnejad2023generic, wei2023human}.
Among these works, HumanMAC \cite{chen2023humanmac} and CoMusion \cite{suncomusion} perform diffusion in input space, relying on a transformer backbone and representing the time dimension in Discrete Cosine Space (DCT), a temporal representation widely employed in SHMP \cite{zhang2021we, mao2021generating, dang2022diverse, wei2023human}.  BeLFusion \cite{barquero2023belfusion} performs latent diffusion \cite{rombach2022highresolution} in a semantically meaningful latent space  but by leveraging a U-Net \cite{dhariwal2021diffusion}. We also wish to perform diffusion in latent space, due to its speed and generalization power \cite{videoworldsimulators2024}.
Differently from deterministic HMP approaches \cite{cui2020learning, li2020dynamic, li2019actional}, stochastic approaches leverage Graph Convolutional Networks (GCN) \cite{kipf2016semi,velivckovic2017graph} on the skeleton graph only at intermediate stages \cite{mao2021generating, dang2022diverse, suncomusion, wei2023human}. %
We build on top of Typed-Graph Convolutions \cite{salzmann2022motron} and design a fully GCN autoencoder and denoising network, retaining the semantic meaning of body joints in latent space and thus embedding an explicit prior on the human skeletal structure in the model architecture.

\subsection{Nonisotropic Probabilistic Diffusion Models}
Diffusion models \cite{ho2020denoising, rombach2022highresolution} usually specify the noisification process through isotropic Gaussian random variables, sampling the noise for each diffusion step following the i.i.d. assumption. Also on manifolds~\cite{luo2022antigen, yim2023se, jagvaral2024unified}, relationships between molecule components are modeled isotropically.
According to recent studies \cite{ge2023preserve, chang2023warped}, the isotropic noise prior may not be the best choice for all tasks: optimizing the noise at inference may improve result quality \cite{eyring2024reno} or solve related tasks \cite{karunratanakul2024optimizing}.
In image generation, few explore non-Gaussian or learned alternatives, by addressing  inverse problems \cite{daras2022soft, stevens2023removing}, or efficiency \cite{zheng2022truncated, lyu2022accelerating}.
When considering nonisotropic processes \cite{kim2022maximum, hoogeboom2022blurring, huang2024blue, voleti2022score}, the generated images are qualitatively comparable but retain longer training and inference time and less scalability \cite{kim2022maximum, hoogeboom2022blurring, huang2024blue}.
We present a novel nonisotropic training formulation by modifying the covariance matrix of the noise addition, making the noisification process aware of joint connections. Since we rely on the known skeleton graph, the covariance matrix is not learned \cite{zheng2022truncated, lyu2022accelerating} but fixed regardless of the input motion. While covariance matrices that depend on the input might not scale well with the problem size \cite{huang2024blue}, our formulation is efficient and comes at no additional computational expenses during both training and inference. 
To the best of our knowledge, we are the first to apply nonisotropic Gaussian diffusion to a structured problem, also showing that our formulation converges with fewer iterations and parameters than its isotropic alternative (see \cref{supp:exp:noniso_vs_iso}).

\begin{figure}

  \centering
  \includegraphics[width=\linewidth]{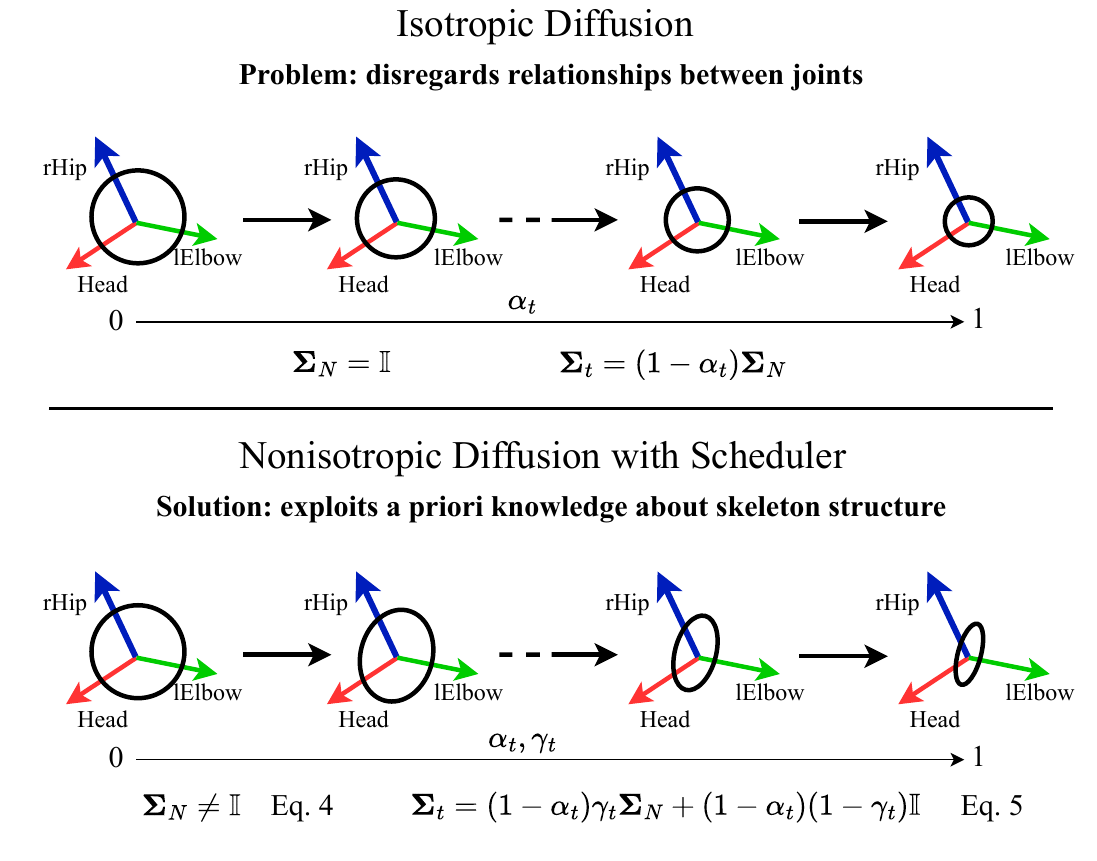}
  \vspace{-20pt}
  \caption{\textbf{Our nonisotropic diffusion formulation.} By diffusing a random variable $\x{}{0} \in \mathbb{R}^{\numjoints}$ where $\numjoints$ is the number of body joints, instead of considering the joint dimensions i.i.d. as in isotropic diffusion, we take into account skeleton connections in $\covadj$. With the scheduler $\gammaT{}{t}$, we design a noise that transitions from isotropic to nonisotropic. Further dimensions can be diffused isotropically.%
  }
  \label{fig:nonisotropic_diffusion}
  \vspace{-0.4cm}
\end{figure}

\vspace{-0.5cm}
\begin{figure*}[ht]
  \centering
  \includegraphics[trim=0cm 0cm 3cm 0cm, clip,width=\textwidth]{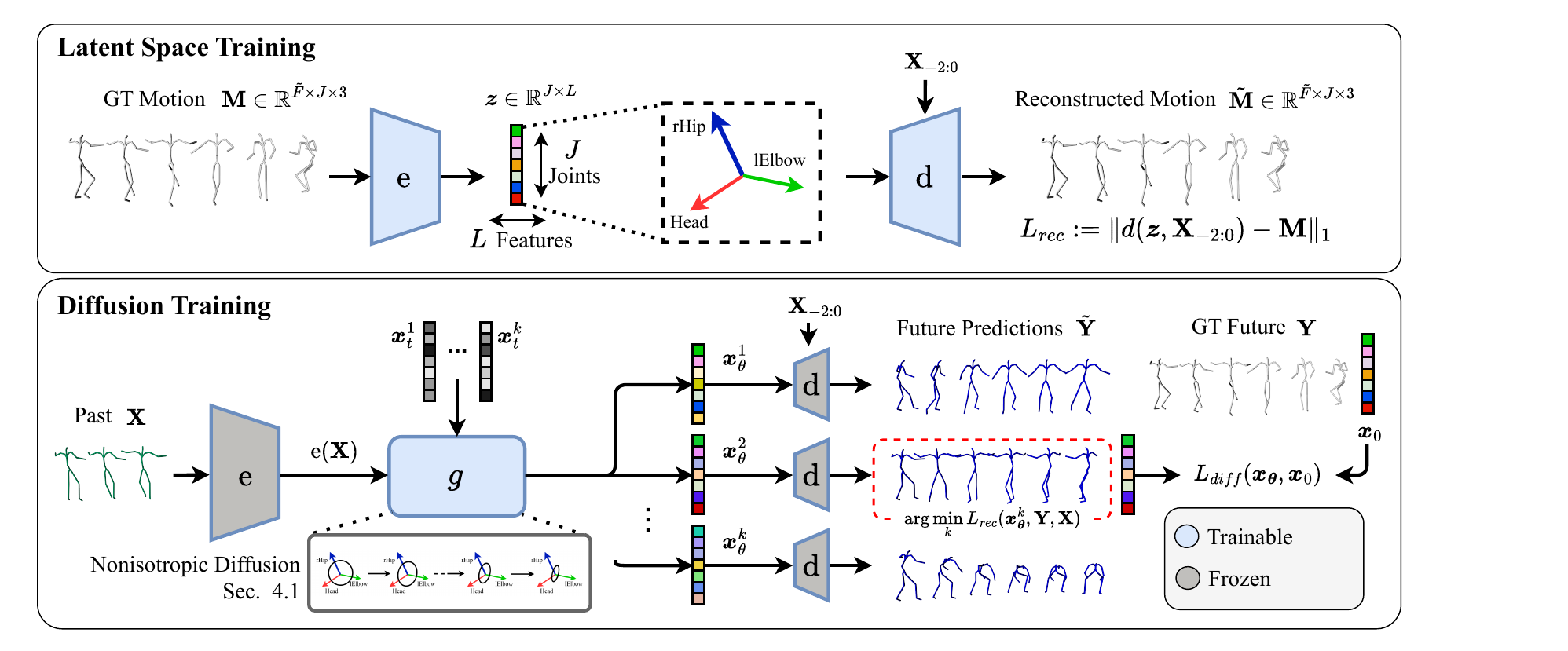}
  \vspace{-0.7cm}
  \caption{\textbf{Overview of \ourmethod}. We first learn a latent space $\boldsymbol{z} \in \mathbb{R}^{J \times L}$ where each of the $J$ latent joint dimensions corresponds to a human body joint, by training encoder $\encnet$ and decoder $\decnet$ to reconstruct human motions sequences. Afterward, the latent joint dimension exhibits correlations similar to human body joints (\cref{method:latent_space}). Here the denoiser network $\diffnet$ conditioned on the past motion $\pastvar{} \in \mathbb{R}^{\pastwindow \times J \times 3}$ is trained via nonisotropic diffusion (\cref{fig:nonisotropic_diffusion} and \cref{method:our-diff}) to generate new latent codes $\x{}{\theta}$. The generated codes are decoded into multiple diverse future motions $\tilde{\futurevar{}}$ matching the past $\pastvar{}$ and the motion closest to the GT is taken to backpropagate the training gradient.}
  \label{fig:skeleton_diffusion}
  \vspace{-0.5cm}
\end{figure*}

\section{Preliminaries}
\vspace{-0.1cm}
\paragraph{Problem Formulation}
\label{sec:method_problem_def}
Human Motion Prediction (HMP) takes as input a past sequence of $\pastwindow$ poses and predicts the corresponding future $\futurewindow$ poses. 
The input motion is defined as $\pastvar{} =  \left[ \pose{}{-\pastwindow+1}, \pose{}{-\pastwindow+2},\dots , \pose{}{0}\right] \in \mathbb{R}^{\pastwindow \times  \numjoints \times 3}$, and the output as $\futurevar{} =  \left[ \pose{}{1},~ \dots , ~ \pose{}{\futurewindow} \right]\in \mathbb{R}^{ \futurewindow \times \numjoints \times 3}$ with $\numjoints$ being the number of human body joints and $\pose{}{\timesteps}$ the 3D body pose at timestep $\timesteps \in \{-\pastwindow+1, -\pastwindow+2, \dots, 0, \dots, F\}$.
Probabilistic HMP considers a set of $\nsamples$ possible future sequences as $\tilde{\futurevar{}}\in \mathbb{R}^{\nsamples \times \futurewindow \times \numjoints \times 3}$ for each observation rather than a single deterministic prediction. 
\vspace{-0.3cm}
\paragraph{Isotropic Gaussian Diffusion}
\label{preliminary:isotropic_diff}
Diffusion generative models aim to learn the distribution $\truedist(\x{}{0})$ of true data samples $\x{}{0}$ by utilizing $T$ unseen hierarchical Markovian latent variables $\x{}{1:T}$ of the same dimensions to define the prior $\prior(\x{}{T})$ and the posterior $\posterior(\x{}{1:T}|\x{}{0})$ distribution:
\vspace{-0.2cm}
\begin{equation}
\truedist(\x{}{0}) = \frac{\truedist(\x{}{0:T})}{\posterior(\x{}{1:T}|\x{}{0})} = \frac{\prior(\x{}{T})\prod_{t=1}^{T}\forward(\x{}{t-1}|\x{}{t})}{ \prod_{t=1}^{T}\posterior(\x{}{t}|\x{}{t-1})}.
\end{equation}
Denoising diffusion probabilistic models \cite{ho2020denoising} define the forward transitions $\forward(\x{}{t}|\x{}{t-1})$ as a linear Gaussian model $\mathcal{N}(\x{}{t};\sqrt{\alphaT{}{t}} \x{}{t-1},  \cov{}{t}) $ with noise scheduler $\alphaT{}{t}$, and the random variables $\x{}{t}$ as i.i.d with isotropic, diagonal covariance 
\vspace{-0.2cm}
\begin{equation}
\label{eq:sigma}
\cov{}{t} = (1 - \alpha_t) \identity.
\end{equation}
The forward process iteratively transforms the true variable $\x{}{0}$ into isotropic Gaussian noise $\prior(\x{}{T}) = \mathcal{N}(\x{}{T}; \boldsymbol{0},\identity)$. The reverse diffusion samples $\x{}{T} \sim  \mathcal{N}(\x{}{T}; \boldsymbol{0},\identity)$ and iteratively applies the denoising transitions $\posterior_\theta(\x{}{t-1}|\x{}{t}) $ parametrized by a neural network $\theta$ to obtain samples from the real data distribution. 

\section{Method} \label{sec:method}
We first present our nonisotropic diffusion formulation (\cref{method:our-diff}), discuss its application in latent space (\cref{method:latent_space}), and then introduce \ourmethod~(\cref{method:skeleton-diffusion}).

\subsection{Nonisotropic Gaussian Diffusion} 
\label{method:our-diff}
Clearly, in HMP every joint position depends on those of its neighbors. Relying on the i.i.d. noise assumption of conventional diffusion models \cite{ho2020denoising, rombach2022highresolution} would overlook such relations. %
Contrary to \textit{isotropic} Gaussian diffusion that denoises all dimensions of a random variable $\x{}{0} \in \mathbb{R}^{\numjoints}$ equally, we propose a \emph{nonisotropic} formulation where each $j$-th dimension  is denoised depending on the kinematic relations of a body with $\numjoints$ joints (\cref{fig:nonisotropic_diffusion}).
\vspace{-0.3cm}
\paragraph{Correlation Matrix $\covadj$}
\label{method:diff:covariance_matrix}
Since joint relationships do not dependend on the diffusion timestep $t$, we define our transition covariance matrix $\cov{}{t}$ in dependence of a \textit{correlation matrix} $\covadj \in \mathbb{R}^{\numjoints \times \numjoints}$ encoding the skeleton structure:
\begin{equation}
\label{eq:purediff}
\cov{}{t} = (1 - \alpha_t) \covadj.
\end{equation}
 A natural choice for $ \covadj$ seems the adjacency matrix $\adjmatrix$ of the simple undirected graph originating from the body skeleton. However, $\adjmatrix$ is an arbitrary matrix not guaranteed to be positive-definite, which is a fundamental property for covariance matrices. Furthermore, to avoid imbalances and exploding values in the noise, the magnitude of $\covadj$ should align with $\identity$. 
 To address these two constraints, we subtract the smallest eigenvalue $\lambda_{\text{min}}(\adjmatrix)$ from the diagonal elements and normalize the result to get the final $\covadj$:
\begin{equation}
\begin{split}
\covadj = \frac{\adjmatrix- \lambda_{\text{min}}(\adjmatrix)\identity}{\lambda_{\text{max}}(\adjmatrix) - \lambda_{\text{min}}(\adjmatrix)} .
\end{split}
\label{eq:sigman}
\end{equation}
We ablate $\adjmatrix$ against two more sophisticated, densely populated choices (\cref{appendix:ablation_sigman}). 
Our formulation comes with negligible computational expenses and can be adapted to any problem that can be defined by an adjacency matrix $\adjmatrix$.
\vspace{-0.7cm}
\paragraph{Nonisotropic Covariance Scheduler}
\label{method:diff:scheduler}
Although the simple solution in \cref{eq:purediff} is already superior to isotropic diffusion (see \cref{results:ablations}), we observe that different diffusion timesteps $t$ relate to different aspects of the generation process. First, the network figures out high-level, global properties of the future motion, and later, fine-grained joints' play a more significant role. With this motivation, we define a noise addition $\cov{}{t}$ that transitions from isotropic to nonisotropic noise: 
\begin{equation}
\label{eq:diff-covt}
    \cov{}{t}= 
    (1 - \alphaT{}{t}) \gammaT{}{t} \covadj  + (1 - \alphaT{}{t}) (1 - \gammaT{}{t} )\identity,
\end{equation}
where $\gammaT{}{t}$ defines a cosine scheduler with opposite behavior to $\alphaT{}{t}$.
Detailed derivation and alternative scheduler formulation explored in early stages are in \cref{appendix:derivation_process}.

\paragraph{Forward and Reverse Nonisotropic Diffusion}
We derive the closed-form $\forward(\x{}{t}|\x{}{0})$ of the forward process as 
\begin{align}
    \x{}{t} & = \sqrt{\alphaT{\bar}{t}}  \x{}{0} + \U\Eigenvalues{\bar}{t}^{1/2}\epsi{}{},
\end{align} 
where the nonisotropic noise is obtained from isotropic noise  $\epsi{}{} \sim \mathcal{N}(\boldsymbol{0}, \textbf{I})$ through the Eigendecomposition of the covariance matrix $ \cov{}{t} = \U \Eigenvalues{}{t} \U^\top$, with $ \Eigenvalues{\bar}{t} =  \gammaT{\tilde}{t} \Eigenvalues{}{t} + (1 - \alphaT{\bar}{t}) \identity $, and $\gammaT{\tilde}{t} = (1 - \alphaT{}{t}) \gammaT{}{t } + \alphaT{}{t} \gammaT{\tilde}{t-1}$. To perform inference, we derive the tractable form for the posterior $\posterior(\x{}{t-1}|\x{}{t})$ as 
\begin{align}
\begin{split}
     \x{}{t-1} = & \mean{}{\posterior} + \U  \Eigenvalues{}{\posterior}\epsi{}{}, \\
     \Eigenvalues{}{\posterior} = & \Eigenvalues{}{t}\Eigenvalues{\bar}{t-1} \Eigenvalues{\bar}{t}^{-1}, \\
    \mean{}{\posterior} = & \sqrt{\alphaT{}{t}} \U\Eigenvalues{\bar}{t}^{-1}\Eigenvalues{\bar}{t - 1} \U^\top\x{}{t} \\
    & + \sqrt{\alphaT{\bar}{t-1}} \U\Eigenvalues{\bar}{t}^{-1}\Eigenvalues{}{t} \U^\top \x{}{0}. \\ 
\label{eq:reverse_diff}
\end{split}
\end{align} 
\vspace{-1.2cm}
\paragraph{Training Objective}
The KL-divergence typically employed to train denoising diffusion models \cite{ho2020denoising, vahdat2021score} can be formulated as Mahalanobis distance. Exploiting the Eigendecomposition, we apply the spectral theorem obtaining
\begin{equation}
\label{eq:loss}
    L_{\mathrm{diff}} (\x{}{\boldsymbol{\theta}}, \x{}{0}, t):= 
    \alphaT{\bar}{t} \|\Eigenvalues{\bar}{t}^{-1/2}\U^\top(\x{}{\boldsymbol{\theta}} - \x{}{0})\|. 
\end{equation}
Detailed derivations are reported in \cref{appendix:kl-loss}, together with the objective for regressing the noise \cite{ho2020denoising} instead of the true variable \cite{rombach2022highresolution}. 
Noticeably, as the eigenvalues $\Eigenvalues{}{N}$ are fixed by the skeleton graph, all required matrices do not depend on the specific input and can be precomputed.

\begin{table*}[t]\footnotesize	
\centering
\setlength{\tabcolsep}{3.5pt}
\begin{tabular}{c  l  rr r r r r r r  r r r r}
\toprule 
\multicolumn{2}{c}{} & \multicolumn{3}{c}{Precision} & \multicolumn{3}{c}{Multimodal GT} & \multicolumn{1}{c}{Diversity} & \multicolumn{1}{c}{Realism} & \multicolumn{4}{c}{Body Realism}\\
 \cmidrule(lr){3-5} \cmidrule(lr){6-8} \cmidrule(lr){9-9} \cmidrule(lr){10-10} \cmidrule(lr){11-14}
\multirow{2}{*}{Type}  & \multirow{2}{*}{Method}
& & & &  & & & &  &\multicolumn{2}{c}{mean $\downarrow$} & \multicolumn{2}{c}{RMSE $\downarrow$} \\
&   
 & ADE $\downarrow$ & FDE $\downarrow$ &  MAE $\downarrow$ & MMADE $\downarrow$ & MMFDE $\downarrow$ & APDE $\downarrow$ & APD $\uparrow$ & CMD $\downarrow$ &  str  & jit & str  & jit\\ %
\midrule

Alg & Zero-Velocity   & 0.755 & 0.992 & 7.779 & 0.814 & 1.015 & 9.292 &  0.000 & 39.262 &  0.00 &  0.00 &  0.00 &  0.00\\
\hline
\multirow{4}{*}{VAE}
& TPK \cite{walker2017pose}  & 0.656 & 0.675 & 10.191 & 0.658 & 0.674 & 2.265 & 9.283  & 17.127 & 7.34 & 0.34 &  9.69 & 0.48 \\ %
& DLow \cite{yuan2020dlow}  & 0.590 & 0.612 & 8.510 & 0.618 & 0.617& 4.243 & {13.170}  & {15.185} & 8.41 & 0.40 & 11.06 & 0.58\\ %
& GSPS \cite{mao2021generating}  & 0.563 & 0.613 & 9.045 & 0.609 & 0.633 & 4.678 & 12.465  & 18.404 & 6.65 & 0.29 & 8.98 & 0.37\\ %
& DivSamp \cite{dang2022diverse}  & 0.564 & 0.647 &  8.027 & 0.623 & 0.667& 15.837 & \textbf{24.724}  & 50.239 & 11.17 & 0.82 & 16.71 & 1.0\\ %
\midrule
\multirow{3}{*}{DM}
& HumanMAC \cite{chen2023humanmac}  & 0.511 & 0.554 & - & 0.593 & 0.591& - & 9.321  & - & - & -& - & -\\
& BeLFusion \cite{barquero2023belfusion}  & {0.513} & {0.560} & 7.125 & {0.569} & {0.585} & \textbf{1.977} & 9.376 & 16.995 & 7.19 & 0.34 & 9.03 & \underline{0.34}\\ %
 & CoMusion \cite{suncomusion} & \underline{0.494} & \underline{0.547} & 
 \underline{6.715} & 
 \textbf{0.469} & \textbf{0.466}& 2.328 & \underline{10.848}  & \textbf{9.636} & \underline{4.04} & \underline{0.25}& \underline{5.63} & 0.52\\ %
 \cmidrule{1-14}
 \morecmidrules\cmidrule{1-14}

DM &  \ourmethod & \bfseries 0.480 & \bfseries 0.545 & \textbf{6.124} & \underline{0.561} & \underline{0.580} & \underline{2.067} & 9.456 & \underline{11.417} & \bfseries 3.15 & \bfseries 0.20 & \bfseries4.45 & \bfseries0.26 \\ %

\bottomrule
\end{tabular}
\vspace{-0.25cm}
\caption{\textbf{Quantitative results on AMASS \cite{mahmood2019amass}}. The best results are highlighted in \textbf{bold}, second-best are \underline{underlined}. The symbol `-' indicates that the results are not reported in the baseline work. We achieve state-of-the-art performance, while the VAE-based method with the highest diversity, DivSamp, displays the worst limb stretching and limb jitter.} 
\label{tab:main_amass}
\vspace{-0.55cm}
\end{table*}

\subsection{Correlated Latent Space}
\label{method:latent_space}

\vspace{-0.1cm}
\paragraph{Extending with i.i.d Features}
While our nonisotropic diffusion has been defined in \cref{method:our-diff} to operate on $\latentvar{} \in \mathbb{R}^{ \numjoints}$, here we extend the formulation to multiple dimensions and opt for a two-dimensional latent representation $\latentvar{} \in \mathbb{R}^{ \numjoints \times \latentdim}$, shown effective in other domains \cite{rombach2022highresolution} but not applied to HMP before. Every $j$-th body joint is described by a feature vector of dimension $\latentdim$, dimension which does not explicitly encode information \underline{between} joints. Thus we assume i.i.d noise over this dimension and diffuse it isotropically, allowing for a richer feature representation. 
\vspace{-0.43cm}
\paragraph{Correlations in Latent Space}
The foundation behind our nonisotropic diffusion formulation of \cref{method:our-diff} is the existing correlation between body joints, described in the adjacency matrix $\adjmatrix$ and reflected in the noisification process by the correlation matrix $\covadj$. In a space where no correlations exist, nonisotropic diffusion is meaningless. For this reason, in our latent space the semantic notion of body joint is intact and the correlations between each $j$-th dimension resembles human body joints motions (see \cref{appendix:skeleton_latent_space} and \cref{fig:LatentSpace_GTembeds}). 
\vspace{-0.15cm}
\subsection{SkeletonDiffusion}
\label{method:skeleton-diffusion}
\vspace{-0.15cm}
SkeletonDiffusion implements our nonisotropic diffusion formulation (\cref{method:our-diff}) in latent space (\cref{method:latent_space}). To obtain an explicit prior on realistic motions, we embed the knowledge about skeletal connections also in the network architecture. 
\vspace{-0.5cm}
\paragraph{Joint-Attentive GCN}
Applying our nonisotropic diffusion in latent space requires retaining the semantic meaning of each body joint (\cref{method:latent_space}). 
We choose a fully GCN architecture and perform graph attention on the skeleton joints via Typed-Graph Convolutions \cite{salzmann2022motron}.
 For each layer taking as input features $\x{}{} \in \mathbb{R}^{ \numjoints \times D_{in}}$, we define a feature extraction matrix $\W{j} \in \mathbb{R}^{ D_{in} \times D_{out}}$ for each joint $j$ with shared weights depending on the specific joint, and a feature aggregation matrix $\mathbf{G} \in \mathbb{R}^{ \numjoints \times \numjoints}$. The features $\feat{} \in \mathbb{R}^{\numjoints \times D_{out}}$ are first extracted for each joint $j$ independently as $\hat{\mathbf{f}}$ and then aggregated through
 \vspace{-0.15cm}
\begin{equation}
\feat{} = \mathbf{G}\cdot \hat{\mathbf{f}},\, \text{with}\,\, \hat{\mathbf{f}}^{j} = \W{j}\cdot \x{}{}^j.
\label{eq:tg-linear}
\end{equation}
We further define multi-head self-attention \cite{vaswani2017attention} on a joint level as Typed-Graph Attention  and chose it as the architecture of the denoiser network.  
Both encoder and decoder are GRUs, exploiting the convenient inductive biases of recurrent neural networks for motion modeling \cite{lyu20223d} (\cref{appendix:graph_arch}). With such architecture, the prior on the body joints is explicitly encoded in every layer.
\vspace{-0.35cm}
\paragraph{Autoencoder and Latent Space Training}
Given an input motion $\motionvar{} = \futurevar{0:\tilde{\futurewindow}} \in \mathbb{R}^{ \tilde{\futurewindow} \times \numjoints \times 3}$ of arbitrary length $\tilde{\futurewindow}\sim  {\mathcal {U}}\{1, \futurewindow\}$, the encoder $\encnet$ compresses the complex temporal information into latent space variables $\latentvar{} = \encnet(\motionvar{}) \in \mathbb{R}^{ \numjoints \times \latentdim}$, where the joint dimension $\numjoints$ is kept intact and the latent dimension $\latentdim$ contains both temporal and spatial information. 
The decoder $\decnet$ learns to reconstruct the latent variable into a motion $\tilde{\motionvar{}} = \decnet (\latentvar{}, \pastvar{-2:0})$. Here, conditioning the decoder on the previous two frames encourages smooth transitions between past and future \cite{barquero2023belfusion}. 
The autoencoder is trained to reconstruct a motion according to the objective:  
 \vspace{-0.05cm}
\begin{equation}
  L_{\mathrm{autoenc}} = L_{\mathrm{rec}} (\encnet(\motionvar{}), \motionvar{}, \pastvar{-2:0}),
  \label{eq:autoenc_loss}
\end{equation}
where the reconstruction loss is defined as
 \vspace{-0.05cm}
\begin{equation}
  L_{\mathrm{rec}} (\latentvar{}, \motionvar{}, \pastvar{-2:0}):= \| \decnet(\latentvar{}, \pastvar{-2:0}) - \motionvar{}\|_1 .
  \label{eq:rec_loss}
\end{equation}
We aim for a strong temporal representation, and let the latent space learn a general motion distribution of arbitrary length, fitting both observation and future motions.
To avoid collapse towards the motion mean of the training data \cite{bhattacharyya2018accurate, wang2021simple}, we employ curricular learning \cite{bengio2009curriculum, wang2021simple, adeli2021tripod}. 
\begin{figure*}[t]
  \centering
    \includegraphics[width=\textwidth]{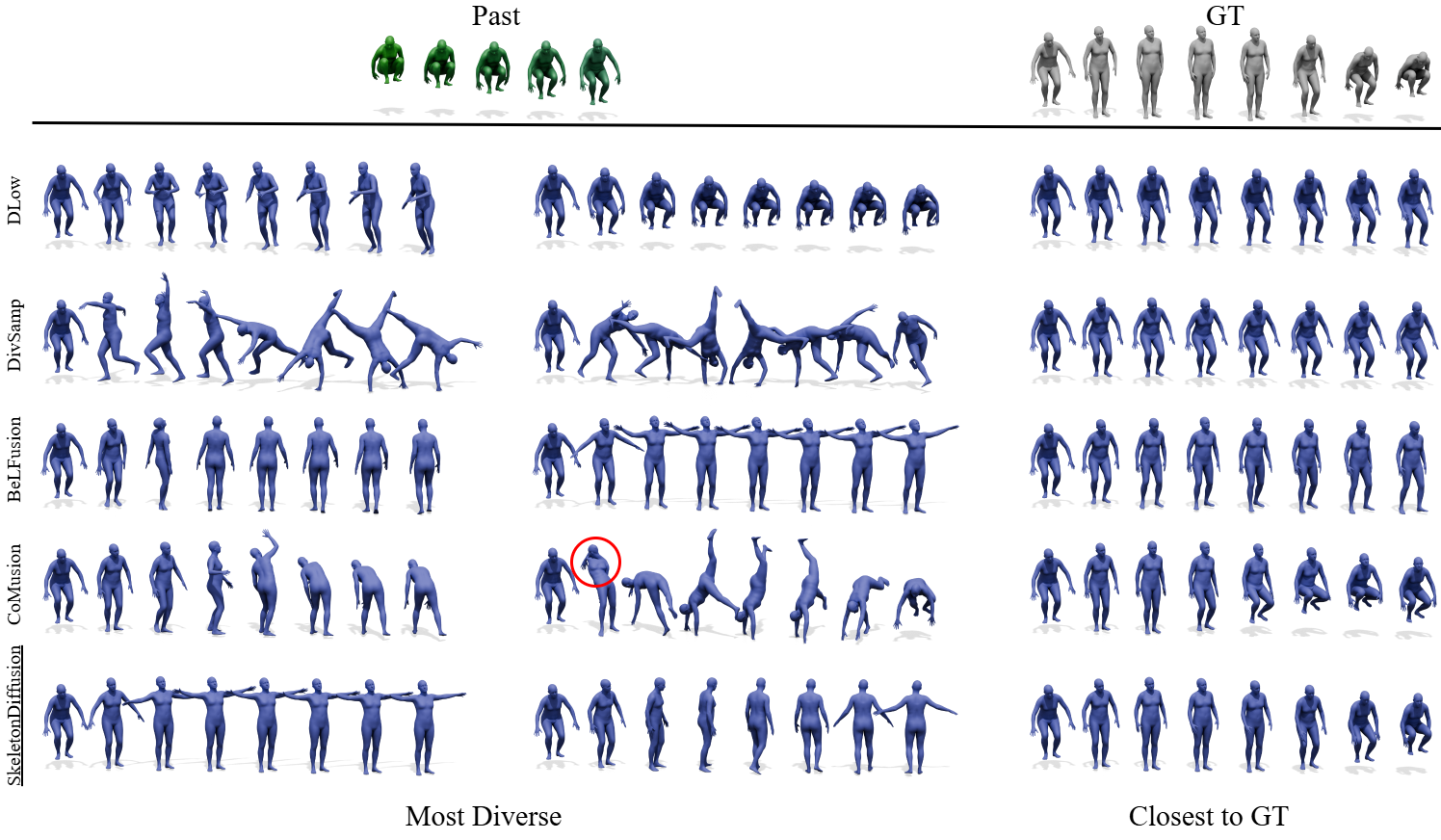}
  \vspace{-0.7cm}
  \caption{\textbf{Qualitative Results on AMASS~\cite{mahmood2019amass}.} On the top, we report the input \textbf{\textcolor{pastgreen}{past}} observation and the \textbf{\textcolor{futurewhite2}{ground truth future}}. The following rows display the corresponding \textbf{\textcolor{bluepred2}{predictions}} for each method: on the right, the closest to GT according to ADE, and on the two left–most columns, the two furthermost. Our closest competitors generate realistic motions but do not include a motion close to the GT (BeLFusion) or present evident unrealistic artifacts (CoMusion).  Our method is the only one to produce \textit{realistic} and \textit{diverse} motions.}
  \label{fig:qualitative}
  \vspace{-0.6cm}
\end{figure*}

\paragraph{Latent Nonisotropic Diffusion} 
In latent space, the denoising network $\diffnet$ learns via our nonisotropic diffusion formulation to denoise true latent variables $\latentvar{}=\encnet(\futurevar{})$ conditioned on past observations $\encnet(\pastvar{})$. 
Instead of predicting the noise $\epsi{}{t}$ \cite{ho2020denoising, rombach2022highresolution}, we directly approximate the true latent code $\x{}{0} :=\latentvar{}$ \cite{ramesh2022hierarchical, barquero2023belfusion} as $\x{}{\boldsymbol{\theta}}= \diffnet(\x{}{t}, \encnet(\pastvar{}), t)$. 
To implicitly enforce diversity \cite{barquero2023belfusion, suncomusion, gupta2018social}, we relax Eq.~\eqref{eq:loss} by sampling $k=50$ predictions at each iteration and backpropagating the gradient only through the sample closest to the GT: 
\vspace{-0.07cm}
\begin{equation}
\hspace{-0.11cm}
\label{eq:loss2}
    L_{\mathrm{gen}}= \mathbb{E}_{\futurevar{}, \pastvar{}, t}  L_{\mathrm{diff}}(\arg \min_k L_{\mathrm{rec}}(\x{}{\boldsymbol{\theta}}^k, \futurevar{}, \pastvar{}), \encnet(\futurevar{}), t)
\end{equation}
Instead of choosing the sample that minimizes the diffusion loss \cite{barquero2023belfusion, suncomusion, gupta2018social}, we choose the prediction that minimizes the reconstruction loss \cref{eq:rec_loss}, finding that this benefits diversity in the generated ensemble (\cref{appendix:ablation_k}). At inference, we denoise multiple latent codes $\x{}{\theta}$ according to our reverse formulation \cref{eq:reverse_diff}. The generated latent codes are then decoded into future predictions $\tilde{\futurevar{}}$.

\vspace{-0.1cm}
\section{Experiments}
\subsection{Experimental Settings}
\paragraph{Baselines} We compare \ourmethod~with state-of-the-art approaches \cite{suncomusion, barquero2023belfusion, chen2023humanmac, dang2022diverse,walker2017pose, yuan2020dlow, mao2021generating} and include the ZeroVelocity baseline, competitive in HMP \cite{martinez2017human, barquero2022comparison} by simply outputting the last seen pose for every future timestep. %
\vspace{-0.8cm}
\paragraph{Datasets}
We evaluate on the large-scale dataset AMASS \cite{mahmood2019amass} according to the cross-dataset evaluation protocol~\cite{barquero2023belfusion, suncomusion}. We aim to test SHMP methods with real-world data obtained not from MoCap but from noise sources (e.g., RGB cameras, and sparse IMUs). To this end, we perform zero-shot experiments on 3D Poses in the Wild (3DPW)~\cite{vonMarcard2018} for models trained on AMASS, and adapt the recent in-the-wild, large-scale dataset FreeMan\cite{Wang2023freeman} to the motion prediction task and retrain on it various state-of-the-art methods. 
We deem the conventionally employed Human3.6M dataset \cite{Ionescu2014} less representative (only 7 subjects) and discuss it directly in \cref{appendix:h36m}. 
As in previous works, we predict the next 2s into the future from observations of 0.5s. %

\vspace{-0.35cm}
\paragraph{Metrics and Body Realism}
\label{metrics:limbs}
Recent SHMP works concentrate on four factors: precision, coverage of the ground truth test distribution (\textit{multimodal} metrics), diversity, and realism. We employ conventional metrics \cite{yuan2020dlow, barquero2023belfusion} and report their definition in \cref{app:metrics}.
While the CMD  metric addresses realism, it is solely expressed in terms of joint velocities. The actual \textit{Body realism}, e.g., bone lengths preservation along the motion, although crucial for meaningful predictions, is overlooked. Even worse, artifacts such as changes in limb lengths over time (limb stretching) and frequent inconsistencies between consecutive frames (limb jitter) \emph{impact other metrics}, for example, by causing more diversity in the predictions, and so higher APD value (further experiments in \cref{appendix:body_realism}). This motivates us to investigate this aspect and propose new metrics. Given a future ground truth sequence $\futurevar{}$ with $\numlimbs$ limbs (or bones) and a predicted sequence $\tilde{\futurevar{}}$, for each frame $\timesteps$ of the prediction associated pose $\predpose{}{\timesteps}$, we denote the length of the $j$-th limb as $\tilde{b}^{j}_{\timesteps} \in \mathbb{R}$. With $b^j_{} \in \mathbb{R}$ being the ground truth length of the $j$-th limb, we define the normalized $j$-th limb length error $e^j_\timesteps$ and limb jitter $v^j_\timesteps$ at a time $\tau$ as:
\begin{equation}
    e^j_\timesteps := \frac{1}{b^j}\left| b^j - \tilde{b}^j_\timesteps \right|, \quad v^j_\timesteps := \frac{1}{b^j} \left| \tilde{b}^j_{\timesteps+1} - \tilde{b}^j_\timesteps \right|.
\end{equation}
By calculating the mean and root mean square error (RMSE) of $e^j_\timesteps$ and $v^j_\timesteps$ over the time dimension, we define four body realism metrics:  \textit{mean} for stretching  \textit{str} and jitter \textit{jit}, and analogously $\text{RMSE}$. We also introduce the mean angle error (MAE) as complementary precision metric.

\begin{table*}[t]\footnotesize	
\centering
\setlength{\tabcolsep}{3.5pt}
\begin{tabular}{lrrrrrrr rrrr}
\toprule  
& \multicolumn{3}{c}{Precision} & \multicolumn{2}{c}{Multimodal GT} & \multicolumn{1}{c}{Diversity} & \multicolumn{1}{c}{Realism} & \multicolumn{4}{c}{Body Realism}\\
  \cmidrule(lr){2-4} \cmidrule(lr){5-6} \cmidrule(lr){8-8} \cmidrule(lr){7-7} \cmidrule(lr){9-12}
 \multirow{2}{*}{Method}  & &  & & & &  & & \multicolumn{2}{c}{mean $\downarrow$} & \multicolumn{2}{c}{RMSE $\downarrow$} \\  
 & ADE $\downarrow$ & FDE $\downarrow$ & MAE $\downarrow$ & MMADE $\downarrow$ & MMFDE $\downarrow$ & APD $\uparrow$ & CMD $\downarrow$ &  str  & jit & str  & jit\\
 
\midrule
Zero-Velocity &  0.603 & 0.835 & 9.841 & 0.687 & 0.865 & 0.000 & 14.734 & 6.37 & 0.00 & 6.37 & 0.00 \\
\hline
HumanMAC \cite{chen2023humanmac}  & 0.415 & 0.511 & 8.630 & 0.537 & 0.600 & 5.426 & \bfseries{2.025} & \underline{7.91}  & 1.49 & 11.89 & 1.84\\
BeLFusion \cite{barquero2023belfusion}  & 0.420 & 0.495 & 8.494 &\bfseries0.496 & \underline{0.516} & 5.209 & 6.306 & 10.46 & \textbf{0.41} & 11.93 & \textbf{0.54}\\
 CoMusion \cite{suncomusion} & 
 \underline{0.389} & \underline{0.480} & \underline{7.812} & 0.527 & 0.525 & \underline{6.687} &  \underline{2.764}  & 7.94 & 0.81 & \underline{10.27} & 1.05\\
\cmidrule{1-12}\morecmidrules\cmidrule{1-12}
 \ourmethod~(\textbf{Ours}) & \bfseries0.374  &  \bfseries0.457 & \textbf{7.424} & \underline{0.506} & \bfseries0.508 & \bfseries 6.732 & 3.166 & \bfseries 7.58  & \underline{0.51}  & \bfseries 9.64  & \underline{0.66}\\ %
\bottomrule
\end{tabular}
\vspace{-0.1cm}
\caption{\textbf{Quantitative results on FreeMan~\cite{Wang2023freeman}}. The best results are highlighted in \textbf{bold}, second best are \underline{underlined}. \ourmethod achieves the best precision and diversity on noisy real-world data while maintaining consistent body realism.} 
\label{tab:main_freeman}
\vspace{-0.4cm}
\end{table*}

\subsection{Results} 
\label{sec:experiments}
\paragraph{Large-scale Evaluation on AMASS}  Following the cross-dataset evaluation protocol \cite{barquero2023belfusion}, we train on a subset of datasets belonging to AMASS and test on others (\cref{tab:main_amass}). Starting from the conventional metrics evaluation, our method already achieves state-of-the-art performance on the majority of the metrics, with a significant improvement on precision. Among other Diffusion-based methods (DM), \ourmethod~and CoMusion contend with each other for first and second place according to diversity, realism, and multimodal metrics. Interestingly, the MAE values for DLow and GSPS do not reflect the performance ranking of the other precision metrics, while instead this holds for DM methods. 
Although VAE-methods tend to show higher diversity values such as of \cite{xu2022diverse, yuan2020dlow}, as already mentioned by previous works \cite{barquero2023belfusion, chen2023humanmac}, these values may often be the consequence of unrealistic motions with irregularities between past and future or inconsistent speed. From the qualitative example reported in \cref{fig:qualitative}, we notice that both the most diverse predictions of DivSamp represent a cartwheeling motion. While such motions may geometrically be diverse from each other and thus increase the APD, they are not only not semantically diverse but also not consistent with the past observation. Instead, our predictions are diverse at no expenses of realism (see also \cref{fig:teaser}).
\vspace{-0.25cm}
\paragraph{Body Realism and Diversity} 
\begin{figure}[t]
  \centering
   \includegraphics[width=0.9\linewidth]{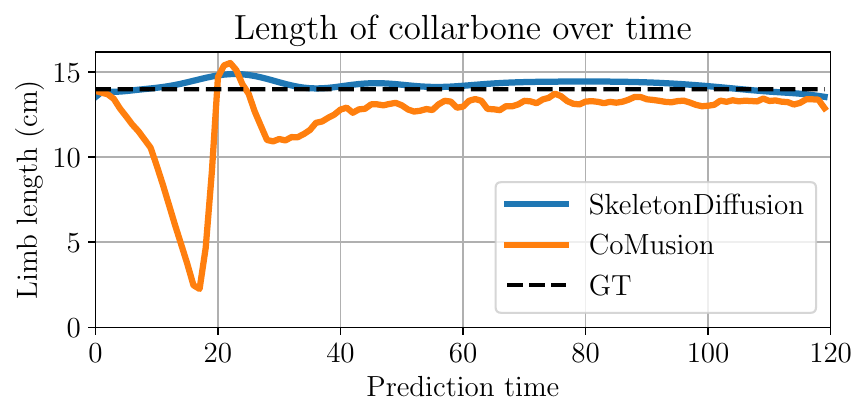}

  \vspace{-0.4cm}
  
   \caption{\textbf{A qualitative example of collarbone bone length evolution} for a single predicted motion from AMASS. \ourmethod~keeps the bone length consistent over time and close to the GT, while CoMusion shows inconsistencies of large magnitude.}
   \label{fig:jitter_comusion}
   \vspace{-0.4cm}
\end{figure}

\begin{figure}[t]
\centering
  \includegraphics[width=0.93\linewidth]{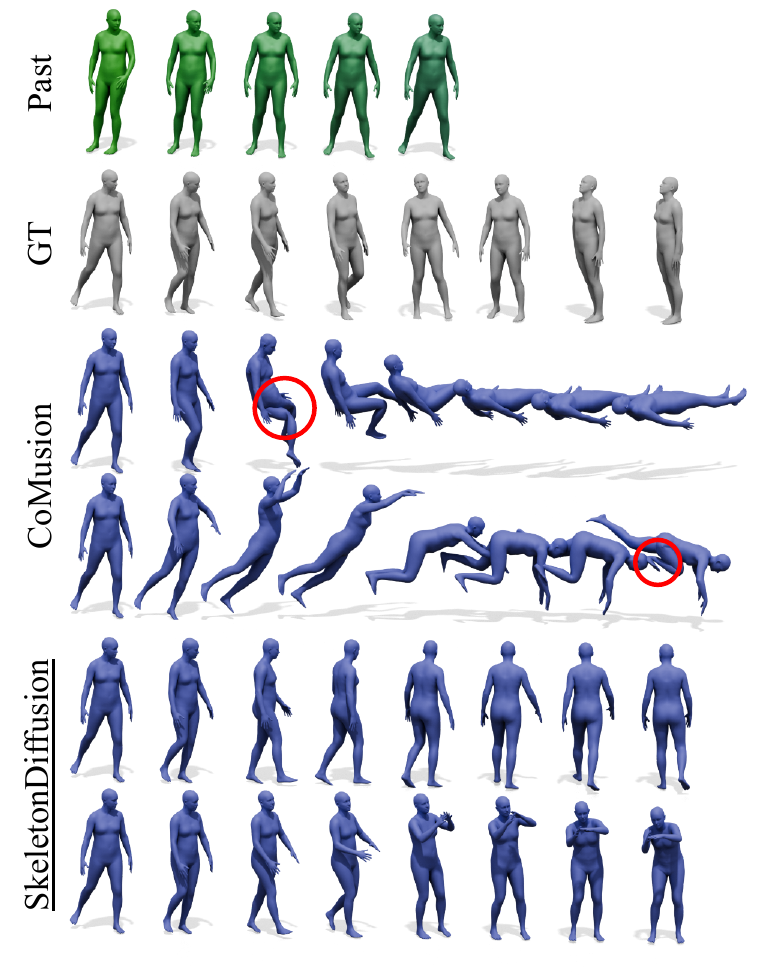}

\vspace{-0.5cm}

\caption{\textbf{Qualitative results of zero-shot on 3DPW~\cite{vonMarcard2018}} for models trained on AMASS~\cite{mahmood2019amass}. CoMusion displays limb twisting, while our predictions are realistic and consistent.}
\label{fig:3dpw-qualitative}
\vspace{-0.4cm}
\end{figure}

\begin{figure}[t]
  \centering
   \includegraphics[width=0.9\linewidth]{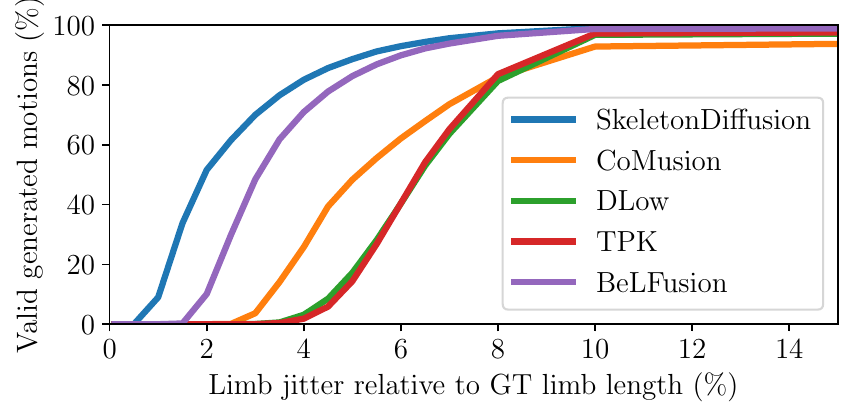}
    \vspace{-0.2cm}
   \caption{\textbf{Motions' validity on different error tolerance on AMASS~\cite{mahmood2019amass}}. For every method, we show the evolution of valid motions quantity (y-axis) for which the maximal error is below a given threshold (x-axis). \methodname~presents consistently the highest number of valid poses. CoMusion and VAE methods cannot generate predictions with an error lower than 2.5\%.}
   \label{fig:valid_samp_jitter}
   \vspace{-0.2cm}
\end{figure}

\begin{table*}[t]\footnotesize	
\centering
\setlength{\tabcolsep}{4.5pt}
\begin{tabular}{c  l  rr r r r r r   r r r r}
\toprule 
\multicolumn{2}{c}{ } & \multicolumn{3}{c}{Precision} & \multicolumn{2}{c}{Multimodal GT} & \multicolumn{1}{c}{Diversity} & \multicolumn{1}{c}{Realism} & \multicolumn{4}{c}{Body Realism}\\
 \cmidrule(lr){3-5} \cmidrule(lr){6-7} \cmidrule(lr){9-9} \cmidrule(lr){8-8} \cmidrule(lr){10-13}
\multirow{2}{*}{Type}  & \multirow{2}{*}{Method} & & & &  & & &  &  \multicolumn{2}{c}{mean $\downarrow$} & \multicolumn{2}{c}{RMSE $\downarrow$} \\
&   & ADE $\downarrow$ & FDE $\downarrow$ & MAE $\downarrow$ & MMADE $\downarrow$ & MMFDE $\downarrow$ & APD $\uparrow$ & CMD $\downarrow$ &  str  & jit & str  & jit\\ %
\midrule
Alg & Zero-Velocity & 0.755 & 1.011 & 7.294 & 0.777 & 1.013 & 0.000 & 40.695 & 0.00 & 0.00 & 0.00 & 0.00 \\
\hline
\multirow{4}{*}{VAE} & TPK \cite{walker2017pose}                   & 0.648 & 0.701 & 9.963 & 0.665 & 0.702 &  9.582 & 13.136 &  7.61 & 0.36 & 10.02 & 0.51 \\
& DLow \cite{yuan2020dlow}                    & 0.581 & 0.649 & 8.820 &  0.602 & 0.651 &\underline{ 13.772} & 11.977 &  8.53 & 0.42 & 11.28 & 0.61 \\ 
& GSPS \cite{mao2021generating}               & 0.552 & 0.650  & 8.469 & 0.578 & 0.653 & 11.809 & 12.722 &  6.38 & 0.29 &  8.65 & 0.35 \\
& DivSamp \cite{dang2022diverse}          & 0.554 & 0.678 & 7.647 &  0.593 & 0.686 & \textbf{24.153} & 46.431 & 11.04 & 0.78 & 16.31 & 1.01 \\
\midrule
\multirow{2}{*}{DM} & BeLFusion \cite{barquero2023belfusion}      & 0.493  & 0.590 & 6.727 & \textbf{0.531 }& 0.599 &  7.740 & 17.725 &  6.47 & \underline{0.22} &  7.96 & \underline{0.29} \\
& CoMusion & \underline{0.477} & \textbf{0.570} & 6.830 & 0.540 &  \textbf{0.587 }& 11.404 & \textbf{7.093} & \underline{4.01} & 0.38 & \underline{5.54} & 0.50 \\
\cmidrule{1-13} \morecmidrules \cmidrule{1-13}
DM & \ourmethod~(\textbf{Ours})                           & \textbf{0.472} &  \underline{0.575} &  6.025 & \underline{0.535} & \underline{0.594} & 9.814 & \underline{10.474} &  \textbf{3.02} & \textbf{0.17} &  \textbf{4.16} & \textbf{0.23} \\

\bottomrule
\end{tabular}
\vspace{-0.1cm}
\caption{\textbf{Zero-Shot evaluation on 3DPW~\cite{vonMarcard2018} for models trained on AMASS~\cite{mahmood2019amass}}. The best results are highlighted in \textbf{bold}, second best are \underline{underlined}. While CoMusion's limb jitter worsens, we present the highest body realism accompanied by solid performance.} 
\label{tab:main_3dpw}
\vspace{-0.3cm}
\end{table*}

On the right-most part of \cref{tab:main_amass}, we analyze limb stretching and jittering in the methods' predictions with our body realism metrics. First, this issue particularly affects  VAE approaches, and the two methods with the highest APD are also the two with the largest errors on all four metrics. This supports our intuition that diversity may benefit from artifacts and inconsistencies. \ourmethod~presents the best metrics by a large margin, highlighting the contribution of our prior on the skeleton structure. CoMusion displays much worse body realism and is the third worst in terms of RMSE error for the jittering.  We  qualitatively visualize the inconsistency of a limb (the collarbone) for a sequence in \cref{fig:jitter_comusion}, reporting its length variation over time. Compared to the ground-truth length (the dashed line), CoMusion shows drastic changes already in the early frames. \ourmethod~is much more consistent over time, remaining quite close to the real length. Finally, we stress the impact of such bone artifacts by considering the case in which an application has a hard requirement about the maximal admitted error for a sequence. Namely, if a sequence faces a bone stretching above a given threshold, it is considered unreliable and so discarded. %
In \cref{fig:valid_samp_jitter}, we report how the number of valid sequences evolves on AMASS in dependence of such threshold, showing that our method is the most robust while CoMusion performs worst among DM models. %
\vspace{-0.3cm}
\paragraph{Noisy Data and FreeMan Dataset} 
We test for the first time SHMP methods on noisy data acquired from an external RGB camera from the FreeMan dataset~\cite{Wang2023freeman}. In this case, GT poses reach a change in limb length up to 5.6cm, compared to close to zero of the AMASS MoCap setting. Our method achieves the best performance in precision and diversity and, at the same time, achieves the lowest limb stretching. This hints that \ourmethod~ has effectively learned basic properties of the human skeletal structure achieving robustness to unprecise data.
We report the evaluation results in \cref{tab:main_freeman}. On the contrary, BelFusion achieves the worst CMD and limb length variation, showing that their bones increase length consistently over the whole prediction. Our findings highlight the informativeness of our four body realism metrics and how our design choices make \ourmethod~ ready also for data sources not previously considered.
\vspace{-0.2cm}
\paragraph{Zero-Shot Generalization on 3DPW} 
We are also interested in evaluating how \methodname~ generalizes to out-of-distribution motions. Hence, we test the methods trained on AMASS on unseen, real-life scenes from 3DPW~\cite{vonMarcard2018} and report results in \cref{tab:main_3dpw}.   
We notice that, while CoMusion's limb jitter between consecutive frames has worsened in the zero-shot setting, our method shows solid results and consistently the best body realism. We report a qualitative example in \cref{fig:3dpw-qualitative}. CoMusion's predictions appear diverse but present low semantic consistency with the input past. Furthermore, both predictions are humanly unfeasible as they present limb twisting or excessive bending.
\vspace{-0.2cm}
\paragraph{Long Term Prediction and Challenging Scenario}
We autoregressively feed generated motions to obtain a forecasting of 5s out of models trained to predict 2s (Appendix \cref{app:tab:long_term}). We also design a challenging scenario testing on zero-shot generalization and noisy input data simultaneously (Appendix \cref{app:tab:3dpw_noisy}). In both settings we maintain the best realism with a significant gap and showcase state-of-the-art precision and diversity: the explicit inductive bias of \ourmethod~ on the human body structure allows our method to preserve the body realism over time and generalize robustingly to noise and actions unseen at train time. 
\paragraph{Ablations} 
\label{results:ablations}
In the Appendix \cref{tab:ablations}, we report the ablations for the main components of \ourmethod~on AMASS. Our TG-Attention layers improve the GCN architecture in the conventional isotropic diffusion paradigm. While the simple nonisotropic variant of \cref{eq:purediff} achieves state-of-the-art performance, our formulation with the scheduler $\gammaT{}{t}$ further improves the metrics and in particular, precision.
Ablation results regarding the choice of connectivity matrix for $\covadj$ and its normalization are reported in \cref{appendix:ablation_sigman}. We also show (\cref{supp:exp:noniso_vs_iso}) that our nonisotropic formulation requires fewer parameters and training epochs than the isotropic one.

\section{Conclusion}
\label{sec:conclusion}

We present \ourmethod, a latent diffusion model with an explicit inductive bias on the human skeleton trained with a novel nonisotropic Gaussian diffusion formulation. We achieve state-of-the-art performance on stochastic HMP by generating motions that are simultaneously realistic and diverse while being robust to limb stretching according to the evaluation metrics.  
\paragraph{Limitations and Future Work} Similar to previous methods, we restrict our experiments to standard human skeletons, without considering fine-grained joints (e.g., fingers, facial expression). Unfortunately, such data are scarce and difficult to capture. While our body realism metrics address previously disregarded aspects, evaluating stochastic HMP and particularly diversity remains an open challenge. 

\paragraph{Acknowledgments}
This work was supported by the ERC Advanced Grant “SIMULACRON” (agreement \#884679), GNI Project “AI4Twinning”, and DFG project CR 250/26-1 “4D YouTube”.
Thanks to Dr. Almut Sophia Koepke, Yuesong Shen and Shenhan Qian for the proofreading and feedback, Lu Sang for the discussion, Jialin Yang for the applications, Stefania Zunino and the whole CVG team for the support.

{
    \small
    \bibliographystyle{ieeenat_fullname}
    \bibliography{main}
}

\clearpage \newpage
\appendix
\maketitlesupplementary
\tableofcontents
\section{Mathematical Derivations of our Nonisotropic Gaussian Diffusion} \label{appendix:derivation_process}

\subsection{Forward Diffusion Process}

As mentioned in the main paper body, the Gaussian forward transitions are defined as: 
\begin{equation}
    q(\x{}{t}|\x{}{t-1})  = \mathcal{N}(\x{}{t}; \sqrt{\alphaT{}{t}}\x{}{t-1}, \U \Eigenvalues{}{t} \U^\top).\label{eq:correlated_qxt_xt-1}
\end{equation}
allowing us to sample from a transition in dependence of isotropic noise $\epsi{}{t}$ as:
\begin{equation}
    \x{}{t} = \sqrt{\alphaT{}{t}}\x{}{t-1} + \U \Eigenvalues{}{t}^{1/2}\epsi{}{t},
\end{equation}
We can further derive the tractable form of the forward transitions $q(\x{}{t}|\x{}{0})$ by recursively applying  $\x{}{t-1} = \sqrt{\alphaT{}{t-1}} \x{}{t-2} + \U \Eigenvalues{}{t-1}^{1/2} \epsi{}{t-1}$:
\begin{equation}
\begin{aligned}
    \x{}{t} = & \sqrt{\alphaT{}{t}} (\sqrt{\alphaT{}{t-1}} \x{}{t-2} + \U \Eigenvalues{}{t-1}^{1/2} \epsi{}{t-1})+ \U \Eigenvalues{}{t}^{1/2} \epsi{}{t} \\
     = & \sqrt{\alphaT{}{t}} (\sqrt{\alphaT{}{t-1}} (\sqrt{\alphaT{}{t-2}} \x{}{t-3} \\
    & + \U \Eigenvalues{}{t-2}^{1/2} \epsi{}{t-2}) + \U \Eigenvalues{}{t-1}^{1/2} \epsi{}{t-1})+ \U \Eigenvalues{}{t}^{1/2} \epsi{}{t} \\
    = & \dots \\
    = & \sqrt{\alphaT{\bar}{t}}  \x{}{0} + \U\Eigenvalues{\bar}{t}^{1/2}\epsi{}{0} \\
    \sim & \mathcal{N}(\sqrt{\alphaT{\bar}{t}} \x{}{0},\U(\Eigenvalues{\bar}{t})\U^T) 
     = \mathcal{N}(\sqrt{\alphaT{\bar}{t}} \x{}{0},\cov{\bar}{t}),
\end{aligned}
\end{equation}
where we exploit the fact that the isotropic noises can be formulated as $\epsi{}{t-1} \sim \mathcal{N}(\boldsymbol{0}, \alphaT{}{t}\U\Eigenvalues{}{t-1}\U^T ) $, $\epsi{}{t} \sim \mathcal{N}(\boldsymbol{0}, \U\Eigenvalues{}{t}\U^T )$ and that the sum of two independent Gaussian random variables is a Gaussian with mean equals the sum of the two means and the variance being the sum of the two variances. We have thus derived the Gaussian form of the tractable forward diffusion process $q(\x{}{t}|\x{}{0}) = \mathcal{N}(\sqrt{\alphaT{\bar}{t}}  \x{}{0},\U\Eigenvalues{\bar}{t}\U^T)$  for
 
\begin{equation}
\Eigenvalues{\bar}{t} =  \gammaT{\tilde}{t} \Eigenvalues{}{\identity} + (1 - \alphaT{\bar}{t}) \identity 
\end{equation}
\begin{equation}
\begin{aligned}
    \gammaT{\tilde}{t} = &\sum_{i=0}^{t}\gammaT{\bar}{t-i} \alphaT{}{t-i}^{-1} \prod_{j=t-i}^{t} \alphaT{}{j} \\
    = & \alphaT{\bar}{t}\sum_{i=0}^{t}\frac{\gammaT{\bar}{t-i} }{\alphaT{\bar}{t-i}} = \gammaT{\bar}{t} + \alphaT{}{t} \gammaT{\tilde}{t-1}\,.
\end{aligned}
\end{equation}

\subsection{Reverse Diffusion Process}
\label{appendix:diff-reverse}
To perform inference, we need to find a tractable form for the posterior $q(\x{}{t-1} | \x{}{t}, \x{}{0})$ in terms of $\x{}{0}$. With the forms of the Gaussian transitions, through Bayes rule 
\begin{equation}
    q(\x{}{t-1} | \x{}{t}, \x{}{0})  = \frac{q(\x{}{t} | \x{}{t-1}, \x{}{0})q(\x{}{t-1} | \x{}{0})}{q(\x{}{t} |  \x{}{0})} 
\end{equation}
we can start the derivation of the posterior $\mathcal{N}(\x{}{t-1};\mean{}q, \cov{}{q}) $ from 
\begin{equation}
    \frac{\mathcal{N}(\x{}{t};\sqrt{\alphaT{}{t}}\x{}{t-1}, \cov{}{t}) \mathcal{N}(\x{}{t-1};\sqrt{\alphaT{\bar}{t-1}}\x{}{0}, \cov{\bar}{t-1})}{\mathcal{N}(\x{}{t};\sqrt{\alphaT{\bar}{t}}\x{}{0}, \cov{\bar}{t})}.
\end{equation}
Differently from the conventional isotropic diffusion derivation, where this and subsequent derivations are carried out for scalar variables thanks to the i.i.d. assumption, our random variables are correlated and we have to deal with vectorial equations. Hence the posterior mean $\mean{}q$ and covariance $\cov{}{q}$ cannot be derived straightforwardly.  

To address this issue, we exploit the eigenvalue decomposition of $\cov{}{t}$ and notice that the orthogonal matrix $\U$ is a linear transformation preserving the inner product of vectors by definition, and that thus the shape of the posterior probability distribution $q(\x{}{t-1} | \x{}{t})$ stays the same in the isometry of the Euclidean space given 
\begin{equation}
    \x{\tilde}{i} = \U^\top \x{}{i}\,.
\end{equation}
This allows us to 'rotate' the posterior distribution $\posterior(\x{}{t-1}|\x{}{t})$ by the transformation $\U$ and carry out the derivation for a distribution $\posterior(\x{\tilde}{t-1}|\x{\tilde}{t})=\mathcal{N}(\x{\tilde}{t-1};\mean{\tilde}{\posterior}, \Eigenvalues{}{\posterior})$ that now has a diagonal covariance matrix $\Eigenvalues{}{\posterior} = \U^\top \cov{}{\posterior}\U$. Now we can handle each dimension independently, since the matrices in the following derivations are diagonal matrices and this allows us to use the commutative property $\Eigenvalues{}{1} \Eigenvalues{}{2} = \Eigenvalues{}{2} \Eigenvalues{}{1}$. 
In the following, we also make use of the observation:
\begin{align}
    \Eigenvalues{\bar}{t} = & \alphaT{}{t} \Eigenvalues{\bar}{t - 1} + \Eigenvalues{}{t} \label{eq:recursive_eigen}
\end{align}
The mean and variance of the posterior can thus be derived in the isometry space as

\begin{align}
    q(\x{\tilde}{t-1} | \x{\tilde}{t}, & \x{\tilde}{0}) \label{eq:diff_final} \\  
    \begin{split}
    \propto \exp{}
        - \frac{1}{2} & \Big[  
        (\x{\tilde}{t} - \sqrt{\alphaT{}{t}} \x{\tilde}{t-1})^\top \Eigenvalues{}{t}^{-1} (\x{\tilde}{t} - \sqrt{\alphaT{}{t}} \x{\tilde}{t-1}) \Big. \\
        + (& \x{\tilde}{t-1} - \sqrt{\alphaT{\bar}{t-1}} \x{\tilde}{0})^\top \Eigenvalues{\bar}{t-1}^{-1} (\x{\tilde}{t-1} - \sqrt{\alphaT{\bar}{t-1}} \x{\tilde}{0}) \\ 
        & \Big. - (\x{\tilde}{t} - \sqrt{\alphaT{\bar}{t}} \x{\tilde}{0})^\top \Eigenvalues{\bar}{t}^{-1} (\x{\tilde}{t} - \sqrt{\alphaT{\bar}{t}} \x{\tilde}{0}) \Big]
    \end{split} \notag \\
    \begin{split}
     = \exp{}
        - \frac{1}{2}  & \Big[  \x{\tilde}{t-1}^\top \alphaT{}{t} \Eigenvalues{}{t}^{-1} \x{\tilde}{t-1} - 2 \x{\tilde}{t-1}^\top \sqrt{\alphaT{}{t}} \Eigenvalues{}{t}^{-1} \x{\tilde}{t} \Big. \\
      &  + \x{\tilde}{t-1}^\top \Eigenvalues{\bar}{t-1}^{-1} \x{\tilde}{t-1} - 2 \x{\tilde}{t-1}^\top \sqrt{\alphaT{\bar}{t-1}} \Eigenvalues{\bar}{t - 1}^{-1} \x{\tilde}{0} \\
     & \Big. + C(\x{\tilde}{t}, \x{\tilde}{0}) \Big]
     \end{split} \notag \\
     \begin{split}
     \propto \exp{}
        - \frac{1}{2} &  \Big[ \x{\tilde}{t-1}^\top \underbrace{(\alphaT{}{t} \Eigenvalues{}{t}^{-1} + \Eigenvalues{\bar}{t-1}^{-1})}_{\Eigenvalues{}{q}^{-1}} \x{\tilde}{t-1} \\ 
        & - 2 \x{\tilde}{t-1}^\top (\sqrt{\alphaT{}{t}} \Eigenvalues{}{t}^{-1} \x{\tilde}{t} + \sqrt{\alphaT{\bar}{t-1}} \Eigenvalues{\bar}{t - 1}^{-1} \x{\tilde}{0} ) \Big]
    \end{split} \notag \\
    \begin{split}
     = \exp{}
        - \frac{1}{2} &  \Big[ \x{\tilde}{t-1}^\top \Eigenvalues{}{q}^{-1} \x{\tilde}{t-1} \notag \\
        & - 2 \x{\tilde}{t-1}^\top (\sqrt{\alphaT{}{t}} \Eigenvalues{}{t}^{-1} \x{\tilde}{t} + \sqrt{\alphaT{\bar}{t-1}} \Eigenvalues{\bar}{t - 1}^{-1} \x{\tilde}{0} ) \big]
    \end{split} \notag \\
    \begin{split}
     = \exp{}
        - \frac{1}{2} &  \Big[ \x{\tilde}{t-1}^\top \Eigenvalues{}{q}^{-1} \x{\tilde}{t-1} \notag \\
         & - 2 \x{\tilde}{t-1}^\top \Eigenvalues{}{q}^{-1} \Eigenvalues{}{q} (\sqrt{\alphaT{}{t}} \Eigenvalues{}{t}^{-1} \x{\tilde}{t} + \sqrt{\alphaT{\bar}{t-1}} \Eigenvalues{\bar}{t - 1}^{-1} \x{\tilde}{0} ) \Big]. 
    \end{split}\notag 
\end{align}
Comparing Eq.\eqref{eq:diff_final} to $\boldsymbol{x}^\top \boldsymbol{\Sigma}^{-1} \boldsymbol{x} - 2 \boldsymbol{x}^\top \boldsymbol{\Sigma}^{-1} \boldsymbol{\mu} + C = (\boldsymbol{x} - \mean{}{})^\top \boldsymbol{\Sigma}^{-1} (\boldsymbol{x} - \mean{}{})$, we can describe the posterior with the following Gaussian form:
\begin{equation}
q(\x{\tilde}{t-1} | \x{\tilde}{t}, \x{\tilde}{0}) = \mathcal{N}(\x{\tilde}{t-1};\mean{\tilde}{q}, \Eigenvalues{}{q})
\end{equation}
\begin{equation}
\begin{split}
    \Eigenvalues{}{q}& =  \left[ \alphaT{}{t} \Eigenvalues{}{t}^{-1} + \Eigenvalues{\bar}{t-1}^{-1} \right]^{-1} \\
    &=  \left[ \alphaT{}{t} \Eigenvalues{\bar}{t-1}\Eigenvalues{\bar}{t-1}^{-1}\Eigenvalues{}{t}^{-1} + \Eigenvalues{}{t}\Eigenvalues{}{t}^{-1}\Eigenvalues{\bar}{t-1}^{-1}\right]^{-1} \\
    &=  \left[ (\alphaT{}{t} \Eigenvalues{\bar}{t-1} + \Eigenvalues{}{t})\Eigenvalues{\bar}{t-1}^{-1}\Eigenvalues{}{t}^{-1}\right]^{-1} \\
    &=  \Eigenvalues{}{t}\Eigenvalues{\bar}{t-1}(\Eigenvalues{}{t} + \alphaT{}{t} \Eigenvalues{\bar}{t-1})^{-1} \quad \\
    &\stackrel{\text{\eqref{eq:recursive_eigen}}}{=}  \Eigenvalues{}{t}\Eigenvalues{\bar}{t-1} \Eigenvalues{\bar}{t}^{-1} ,
\end{split}
\end{equation}

\begin{align}
    & \mean{\tilde}{q} = \Eigenvalues{}{q} (\sqrt{\alphaT{}{t}} \Eigenvalues{}{t}^{-1} \x{\tilde}{t} + \sqrt{\alphaT{\bar}{t-1}} \Eigenvalues{\bar}{t - 1}^{-1} \x{\tilde}{0} ) \\ 
    & = \Eigenvalues{}{t}\Eigenvalues{\bar}{t-1} (\Eigenvalues{}{t} + \alphaT{}{t} \Eigenvalues{\bar}{t-1})^{-1} (\sqrt{\alphaT{}{t}} \Eigenvalues{}{t}^{-1} \x{\tilde}{t} + \sqrt{\alphaT{\bar}{t-1}} \Eigenvalues{\bar}{t - 1}^{-1} \x{\tilde}{0}) \notag \\ 
    & = \Eigenvalues{\bar}{t}^{-1}(\sqrt{\alphaT{}{t}} \Eigenvalues{\bar}{t - 1} \x{\tilde}{t} + \sqrt{\alphaT{\bar}{t-1}} \Eigenvalues{}{t} \x{\tilde}{0} ) \notag
\end{align}

To obtain the previous definition of $\mean{\tilde}{q}$ and $ \Eigenvalues{}{q}$, we make use of the following equalities, that coincide wih our intuition and understanding of denoising diffusion processes and are reported for completeness:
\begin{align}
    \gammaT{\tilde}{t} = & \gammaT{\bar}{t} + \alphaT{}{t} \gammaT{\tilde}{t-1} \notag \\
    = & \gammaT{\bar}{t} + \alphaT{}{t} \sum_{i=0}^{t-1}\gammaT{\bar}{t-1-i} \alphaT{}{t-1-i}^{-1} \prod_{j=t-1-i}^{t-1} \alphaT{}{j} \notag \\
    = & \sum_{i=-1}^{-1}\gammaT{\bar}{t-1-i} \alphaT{}{t-1-i}^{-1} \prod_{j=t-1-i}^{t} \alphaT{}{j} \notag \\
    & + \sum_{i=0}^{t-1}\gammaT{\bar}{t-1-i} \alphaT{}{t-1-i}^{-1} \prod_{j=t-1-i}^{t} \alphaT{}{j} \\
    = & \sum_{i=-1}^{t-1}\gammaT{\bar}{t-1-i} \alphaT{}{t-1-i}^{-1} \prod_{j=t-1-i}^{t} \alphaT{}{j} \,\, | \,\, \begin{array}{l}
        \text{shift the $i$ index} \\ 
        \text{by 1 ($i:=i + 1$)}
        \end{array} \notag \\ 
    = & \sum_{i=0}^{t}\gammaT{\bar}{t-i} \alphaT{}{t-i}^{-1} \prod_{j=t-i}^{t} \alphaT{}{j} \notag 
\end{align}

\begin{equation}
    \begin{split}
    \Eigenvalues{\bar}{t} = & \alphaT{}{t} \Eigenvalues{\bar}{t - 1} + \Eigenvalues{}{t} \\
    = & \alphaT{}{t} \left( \gammaT{\tilde}{t-1} \Eigenvalues{}{\identity} + (1 - \alphaT{\bar}{t-1}) \identity  \right) + \left(\gammaT{\bar}{t} \Eigenvalues{}{\identity} + (1 - \alphaT{}{t}) \identity \right) \\
    = & \left( \alphaT{}{t}  \gammaT{\tilde}{t-1} + \gammaT{\bar}{t}  \right) \Eigenvalues{}{\identity} + \left(\alphaT{}{t}(1 - \alphaT{\bar}{t-1})+ (1 - \alphaT{}{t}) \right)  \identity \\
    = &  \gammaT{\tilde}{t} \Eigenvalues{}{\identity} + \left(1 - \alphaT{\bar}{t} \right)  \identity \\
    \end{split}
\end{equation}
\begin{equation}
    \cov{\bar}{t} = \alphaT{}{t} \cov{\bar}{t - 1} + \cov{}{t} \label{eq:recursive_cov} \\ 
\end{equation}

We detail how to transform the new mean and covariance into the original coordinate system:
\begin{equation}
\begin{split}
    \cov{}{q} = & \U \Eigenvalues{}{q} \U^\top \\
    = & \U \Eigenvalues{}{t}\Eigenvalues{\bar}{t-1} \Eigenvalues{\bar}{t}^{-1} \U^\top \\
    = & \U \Eigenvalues{}{t} \underbrace{\U^\top \U}_{\identity} \Eigenvalues{\bar}{t-1} \U^\top \U \Eigenvalues{\bar}{t}^{-1} \U^\top \\
    = & \cov{}{t} \cov{\bar}{t-1} \cov{\bar}{t}^{-1},
\end{split}
\end{equation}
\begin{equation}
\begin{split}
    \mean{}{q} = & \U \mean{\tilde}{q} \\
    = & \U \Eigenvalues{\bar}{t}^{-1}(\sqrt{\alphaT{}{t}} \Eigenvalues{\bar}{t - 1} \x{\tilde}{t} + \sqrt{\alphaT{\bar}{t-1}} \Eigenvalues{}{t} \x{\tilde}{0} ) \\
    = & \U \Eigenvalues{\bar}{t}^{-1} \U^\top \U (\sqrt{\alphaT{}{t}} \Eigenvalues{\bar}{t - 1} \U^\top \U \x{\tilde}{t} + \sqrt{\alphaT{\bar}{t-1}} \Eigenvalues{}{t} \U^\top \U \x{\tilde}{0} ) \\
    = & \cov{\bar}{t}^{-1} (\sqrt{\alphaT{}{t}} \cov{\bar}{t - 1} \x{}{t} + \sqrt{\alphaT{\bar}{t-1}} \cov{}{t} \x{}{0} )
\end{split}
\end{equation}

\subsection{Training objective}
\label{appendix:kl-loss}
Denoising diffusion probabilistic models \cite{ho2020denoising} are trained by minimizing the negative log likelihood of the evidence lower bound, which can be simplified  to the KL divergence between the posterior $\posterior(\x{}{t-1} | \x{}{t}, \x{}{0})$ and the learned reverse process $p_{\boldsymbol{\theta}}(\x{}{t-1} | \x{}{t})$. 
Since the covariance matrix is independent of $\theta$, the KL-divergence can be expressed as Mahalanobis distance
\begin{equation}
\begin{split}
    \arg \min_{\boldsymbol{\theta}} D_{\text{KL}}(q(\x{}{t-1} | \x{}{t}, \x{}{0}) \| p_{\boldsymbol{\theta}}(\x{}{t-1} | \x{}{t})) \\ = \arg \min_{\boldsymbol{\theta}} \frac{1}{2} \left[ (\mean{}{\boldsymbol{\theta}} - \mean{}{q})^\top \cov{}{q}^{-1}  (\mean{}{\boldsymbol{\theta}} - \mean{}{q}) \right].
\end{split}
\end{equation}

\paragraph{Regressing the true latent $\x{}{0}$}
We compute the KL divergence in the isometry space with diagonal covariances as
\begin{equation}
\begin{split}
    & \arg \min_{\boldsymbol{\theta}} D_{\text{KL}}(q(\x{\tilde}{t-1} | \x{\tilde}{t}, \x{\tilde}{0}) \| p_{\boldsymbol{\theta}}(\x{\tilde}{t-1} | \x{\tilde}{t})) \\
    = & \arg \min_{\boldsymbol{\theta}} \frac{1}{2} \left[ (\mean{\tilde}{\boldsymbol{\theta}} - \mean{\tilde}{q})^\top \Eigenvalues{}{q}^{-1} (\mean{\tilde}{\boldsymbol{\theta}} - \mean{\tilde}{q}) \right] \\
    = & \left[ \Eigenvalues{\bar}{t}^{-1} \sqrt{\alphaT{\bar}{t-1}} \Eigenvalues{}{t} (\x{\tilde}{\boldsymbol{\theta}} -  \x{\tilde}{0}) \right]^\top \Eigenvalues{}{q}^{-1} \left[ \Eigenvalues{\bar}{t}^{-1} \sqrt{\alphaT{\bar}{t-1}} \Eigenvalues{}{t} (\x{\tilde}{\boldsymbol{\theta}} -  \x{\tilde}{0}) \right] \\
    = & \left[(\x{\tilde}{\boldsymbol{\theta}} -  \x{\tilde}{0}) \right]^\top \alphaT{\bar}{t-1}\Eigenvalues{}{q}^{-1}\Eigenvalues{\bar}{t}^{-2}  \Eigenvalues{}{t}^2 \left[ \Eigenvalues{\bar}{t}^{-1} \sqrt{\alphaT{\bar}{t-1}} \Eigenvalues{}{t} (\x{\tilde}{\boldsymbol{\theta}} -  \x{\tilde}{0}) \right] \\
    = & \left[\x{\tilde}{\boldsymbol{\theta}} -  \x{\tilde}{0} \right]^\top \alphaT{\bar}{t-1}\Eigenvalues{}{t}^{-1}\Eigenvalues{\bar}{t-1}^{-1} \Eigenvalues{\bar}{t} \Eigenvalues{\bar}{t}^{-2}  \Eigenvalues{}{t}^2 \left[\x{\tilde}{\boldsymbol{\theta}} -  \x{\tilde}{0} \right] \\
    = & \left[\x{\tilde}{\boldsymbol{\theta}} -  \x{\tilde}{0} \right]^\top \alphaT{\bar}{t-1}\Eigenvalues{\bar}{t-1}^{-1}  \Eigenvalues{\bar}{t}^{-1}  \Eigenvalues{}{t} \left[\x{\tilde}{\boldsymbol{\theta}} -  \x{\tilde}{0} \right] \\
    = & \left[\x{\tilde}{\boldsymbol{\theta}} -  \x{\tilde}{0} \right]^\top \Eigenvalues{\bar}{t-1}^{-1}  \Eigenvalues{\bar}{t}^{-1}  \alphaT{\bar}{t-1}(\Eigenvalues{\bar}{t} - \alphaT{}{t}\Eigenvalues{\bar}{t-1}) \left[\x{\tilde}{\boldsymbol{\theta}} -  \x{\tilde}{0} \right] \\
    = & \left[\x{\tilde}{\boldsymbol{\theta}} -  \x{\tilde}{0} \right]^\top \Eigenvalues{\bar}{t-1}^{-1}  \Eigenvalues{\bar}{t}^{-1}  (\alphaT{\bar}{t-1} \Eigenvalues{\bar}{t} - \alphaT{\bar}{t}\Eigenvalues{\bar}{t-1}) \left[\x{\tilde}{\boldsymbol{\theta}} -  \x{\tilde}{0} \right] \\
    = & \left[\x{\tilde}{\boldsymbol{\theta}} -  \x{\tilde}{0} \right]^\top (\alphaT{\bar}{t-1} \Eigenvalues{\bar}{t-1}^{-1} - \alphaT{\bar}{t}\Eigenvalues{\bar}{t}^{-1}) \left[\x{\tilde}{\boldsymbol{\theta}} -  \x{\tilde}{0} \right] \\
    = & \left[\x{\tilde}{\boldsymbol{\theta}} -  \x{\tilde}{0} \right]^\top (\|\mean{\tilde}{t-1}\|^2 \Eigenvalues{\bar}{t-1}^{-1} - \|\mean{\tilde}{t}\|^2\Eigenvalues{\bar}{t}^{-1}) \left[\x{\tilde}{\boldsymbol{\theta}} -  \x{\tilde}{0} \right] \\
    = & \left[\x{\tilde}{\boldsymbol{\theta}} -  \x{\tilde}{0} \right]^\top (\tilde{\text{SNR}}(t-1) - \tilde{\text{SNR}}(t)) \left[\x{\tilde}{\boldsymbol{\theta}} -  \x{\tilde}{0} \right] \\
    = &  \|\x{\tilde}{\boldsymbol{\theta}} -  \x{\tilde}{0}\|^2_{(\tilde{\text{SNR}}(t-1) - \tilde{\text{SNR}}(t))^{-1} } \\
    = &  \|\x{\tilde}{\boldsymbol{\theta}} -  \x{\tilde}{0}\|^2_{\boldsymbol{S}^{-1} }
\end{split}
\end{equation}
where we employ the definition of $\tilde{\text{SNR}}(t)) = \|\mean{\tilde}{t}\|^2\Eigenvalues{\bar}{t}^{-1}$ for the signal-to-noise ratio.
The last line denotes the Mahalanobis distance between $\x{\tilde}{\boldsymbol{\theta}}$ and $\x{\tilde}{0}$ with respect to a probability distribution with symmetric positive-definite covariance matrix $ \boldsymbol{S} =(\tilde{\text{SNR}}(t-1) - \tilde{\text{SNR}}(t))^{-1}$.%

As in conventional diffusion training \cite{ho2020denoising}, we train directly with $ \boldsymbol{S} = (\tilde{\text{SNR}}(t))^{-1} $, which in our case translates to $ \boldsymbol{S} ^{-1}= \alphaT{\bar}{t}\Eigenvalues{\bar}{t}^{-1}$ . According to the spectral theorem, for every positive-definite matrix $\boldsymbol{A}$ it holds $\boldsymbol{A}^{-1} = \boldsymbol{W}^\top\boldsymbol{W}$. Since $S$ is diagonal, the spectral theorem translates to $\boldsymbol{S}^{-1} = \boldsymbol{S}^{-1/2\top}\boldsymbol{S}^{-1/2}=\alphaT{\bar}{t}\Eigenvalues{\bar}{t}^{-1}$ with $\boldsymbol{W}^\top :=\boldsymbol{S}^{-1/2} = \sqrt{\alphaT{\bar}{t}}\Eigenvalues{\bar}{t}^{-1/2} = \boldsymbol{W}$ and the  Mahalanobis distance becomes 
\begin{equation}
\begin{split}
    & \arg \min_{\boldsymbol{\theta}} D_{\text{KL}}(q(\x{\tilde}{t-1} | \x{\tilde}{t}, \x{\tilde}{0}) \| p_{\boldsymbol{\theta}}(\x{\tilde}{t-1} | \x{\tilde}{t})) \\
    & = \|\boldsymbol{W}(\x{\tilde}{\boldsymbol{\theta}} -  \x{\tilde}{0})\|^2 %
     = \alphaT{\bar}{t} \|\Eigenvalues{\bar}{t}^{-1/2}(\x{\tilde}{\boldsymbol{\theta}} -  \x{\tilde}{0})\|^2 
\end{split}
\end{equation}
Thus in the original coordinate system the final training objective can be defined as
\begin{equation}
\begin{split}
    & \arg \min_{\boldsymbol{\theta}} D_{\text{KL}}(q(\x{}{t-1} | \x{}{t}, \x{}{0}) \| p_{\boldsymbol{\theta}}(\x{}{t-1} | \x{}{t})) \\
    & = \alphaT{\bar}{t} \|\Eigenvalues{\bar}{t}^{-1/2}\U^\top(\x{}{\boldsymbol{\theta}} - \x{}{0})\|^2 
\end{split}
\end{equation}

\paragraph{Regressing the noise $ \epsi{}{\boldsymbol{\theta}}$}
We report here the necessary equations for regressing the noise $ \epsi{}{\boldsymbol{\theta}}$ instead of the true latent variable  $\x{}{0}$. By applying the reparameterization trick in the isometry space we define
\begin{equation}
        \x{\tilde}{0} = \frac{1}{\sqrt{\alphaT{\bar}{t}}} (\x{\tilde}{t} - \Eigenvalues{}{t}^{1/2} \epsi{}{0})
\end{equation}
By regressing the noise and considering the previous formulation we derive the KL-divergence with an analogous procedure.
\begin{equation}
\begin{split}
    & \arg \min_{\boldsymbol{\theta}} D_{\text{KL}}(q(\x{\tilde}{t-1} | \x{\tilde}{t}, \x{\tilde}{0}) \| p_{\boldsymbol{\theta}}(\x{\tilde}{t-1} | \x{\tilde}{t})) \\
     & = \left[\epsi{}{0} - \epsi{}{\boldsymbol{\theta}}\right]^\top \frac{ \Eigenvalues{}{t}}{\alphaT{\bar}{t}}  (\tilde{\text{SNR}}(t-1) - \tilde{\text{SNR}}(t)) \left[\epsi{}{0} - \epsi{}{\boldsymbol{\theta}} \right] \\
\end{split}
\end{equation}
The training objective in the original covariance space is given by
\begin{equation}
\label{eq:loss-noise}
\begin{split}
    & \arg \min_{\boldsymbol{\theta}} D_{\text{KL}}(q(\x{}{t-1} | \x{}{t}, \x{}{0}) \| p_{\boldsymbol{\theta}}(\x{}{t-1} | \x{}{t})) \\
    = & \left[\epsi{}{0} - \epsi{}{\boldsymbol{\theta}}\right]^\top \frac{\cov{}{t}}{\alphaT{\bar}{t}}  (\text{SNR}(t-1) - \text{SNR}(t)) \left[\epsi{}{0} - \epsi{}{\boldsymbol{\theta}} \right] \\
\end{split}
\end{equation}

\subsection{Alternative Nonisotropic Formulations of $\cov{}{t}$}\label{appendix:sigma_t}

\label{appendix:diff_diffusion}
In this section, we present formulations of the covariance of the forward noising transitions $\forward(\x{}{t}|\x{}{t-1}) = \mathcal{N}(\x{}{t};\sqrt{\alphaT{}{t}} \x{}{t-1},  \cov{}{t}) $ alternative to our nonisotropic formulation with scheduler $\gammaT{}{t}$ defined in \cref{eq:diff-covt}. We report these alternative formulations either because we ablate against them, or because these were discarded in early research stages. Note that for all formulations, the derivation of the tractable forward and posterior still holds, just for a different choice of $\Eigenvalues{\bar}{t}$. 

\subsubsection{Scheduler $\gammaT{}{t}=1$}
\label{appendix:pure-ani}
The most straightforward case of nonisotropic Gaussian diffusion can be obtained by setting $\gammaT{}{t}=1$ in our \cref{eq:diff-covt} 
\begin{equation}
    \cov{}{t} = (1 - \alpha_t) \cov{}{N} = \U (1 - \alpha_t) \Eigenvalues{}{N} \U^\top \,,
\end{equation}
\begin{align}
    \Eigenvalues{}{t} = & (1 - \alphaT{}{t}) \Eigenvalues{}{N}
\end{align}
resulting in nonisotropic noise sampling for the last hierarchical latent $t=T$. We highlight that this choice of $\cov{}{t}$ corresponds to performing conventional isotropic diffusion ($\cov{}{t} = \identity$)  in a normalized space where the dimensions are not correlated anymore (for example through an affine transformation disentangling the joint dimensions, or layer normalization) and transform back the diffused features to the skeleton latent space.

For the tractable form of the forward process $\forward(\x{}{t}|\x{}{0}) = \mathcal{N}(\sqrt{\alphaT{\bar}{t}} \x{}{0},\U\Eigenvalues{\bar}{t}\U^T)$ it follows 
\begin{align}
    \Eigenvalues{\bar}{t} = & (1 - \alphaT{\bar}{t}) \Eigenvalues{}{N}
\end{align}
The computation of the corresponding posterior exploits the following equality:
\begin{equation}
    \begin{split}
    \Eigenvalues{\bar}{t} = & \alphaT{}{t} \Eigenvalues{\bar}{t - 1} + \Eigenvalues{}{t} \\
    = & \alphaT{}{t} (1 - \alphaT{\bar}{t - 1}) \Eigenvalues{}{N} + (1 - \alpha_t) \Eigenvalues{}{N} \\
    = & (\alphaT{}{t} (1 - \alphaT{\bar}{t - 1}) + (1 - \alpha_t)) \Eigenvalues{}{N} \\
    = & (\alphaT{}{t} - \alphaT{}{t}\alphaT{\bar}{t - 1} + 1 - \alpha_t) \Eigenvalues{}{N} \\
    = & (1 - \alphaT{\bar}{t}) \Eigenvalues{}{N} \\
    \end{split}
\end{equation}

\subsubsection{Discarded Scheduler Formulation}
\label{appendix:ani-iso-diff}
As a preliminary study of our correlated diffusion approach, we explored the following covariance:
\begin{equation}
\cov{}{t} = \covadj \alphaT{}{t} + \identity (1 - \alphaT{}{t} )
\end{equation}
\begin{equation}
    \Eigenvalues{}{N}  =  \Eigenvalues{}{N} \alphaT{}{t} +  (1 - \alphaT{}{t} ) \identity
\end{equation} 
As $\cov{}{t} \rightarrow \identity$ for $t \rightarrow T$, we have an identity covariance matrix in the final timestep. 
Adding large quantities of non-isotropic noise in early diffusion timesteps as described did not yield satisfactory results during experiments.
Hence this formulation was discarded at an early research stage. For completeness, we report the covariances of the tractable forward transition as

\begin{equation}
    \Eigenvalues{\bar}{t} = \alphaT{\tilde}{t}\Eigenvalues{}{N}+  (1 - \alphaT{\bar}{t})\identity
\end{equation}
where
\begin{equation}
    \alphaT{\tilde}{t} = \sum_{i=0}^{t}\prod_{j=t-i}^t \alphaT{}{j} = \alphaT{}{t} (1 + \alphaT{\tilde}{t-1}) \label{eq:alpha_tilde}\,.
\end{equation}

\section{Network architecture}
\label{appendix:graph_arch}
\ourmethod's architecture builds on top of  Typed-Graph (TG) convolutions \cite{salzmann2022motron}, a type of graph convolutions designed particularly for human motion prediction.  The conditional autoencoder consists of two shallow TG GRU \cite{salzmann2022motron}. To obtain a strong temporal representation of arbitrary length, thus fitting both observation and ground truth future, we pass the encoder's last GRU state to a TG convolutional layer \cite{salzmann2022motron}. The denoiser network consists of a custom architecture of stacked residual blocks of TG convolutions and TG Attention layers.  Details are available through the code implementation. 
\paragraph{Typed Graph Attention}
\label{appendix:tg-attention}
We introduce Typed Graph Attention (TG Attention) as multi head self-attention deployed through TG convolutions \cite{salzmann2022motron}. To compute scaled dot-product attention as defined by Vaswani et al. \cite{vaswani2017attention} with a scaling factor $d_k$
\begin{equation}
  \mathrm{Attention}(\mathbf{Q}, \mathbf{K}, \mathbf{V}) = \mathrm{softmax}(\frac{\mathbf{Q}\mathbf{K}^T}{\sqrt{d_k}})\mathbf{V}
\end{equation}
 we define the query, key, and value matrices $\mathbf{Q}_i, \mathbf{K}_i, \mathbf{V}_i \in \mathbb{R}^{\numjoints \times D_{out}}$ for each head $i$ with input $\x{}{} \in \mathbb{R}^{\numjoints \times D_{in}}$:
\begin{equation}
  \mathbf{Q}_i = \tglinear{}(\mathrm{RMS} (\x{}{})), \mathbf{K}_i = \tglinear{}(\mathrm{RMS} (\x{}{})), \mathbf{V}_i = \tglinear{}(\mathrm{RMS} (\x{}{})),
  \label{eq:tg-attention}
\end{equation}
where $\tglinear{}$ denotes the TG convolution operation described in \cref{eq:tg-linear} and $\mathrm{RMS}$ the Root Mean Square Norm (RMS)\cite{zhang2019root}, acting as a regularization technique increasing the re-scaling invariance of the model \cite{zhang2019root,vaswani2017attention}.

\begin{figure}[t]
  \centering
    \vspace{0.0cm}
    \includegraphics[trim={0cm, 1cm, 0cm, 0cm},clip, width=0.9\textwidth]{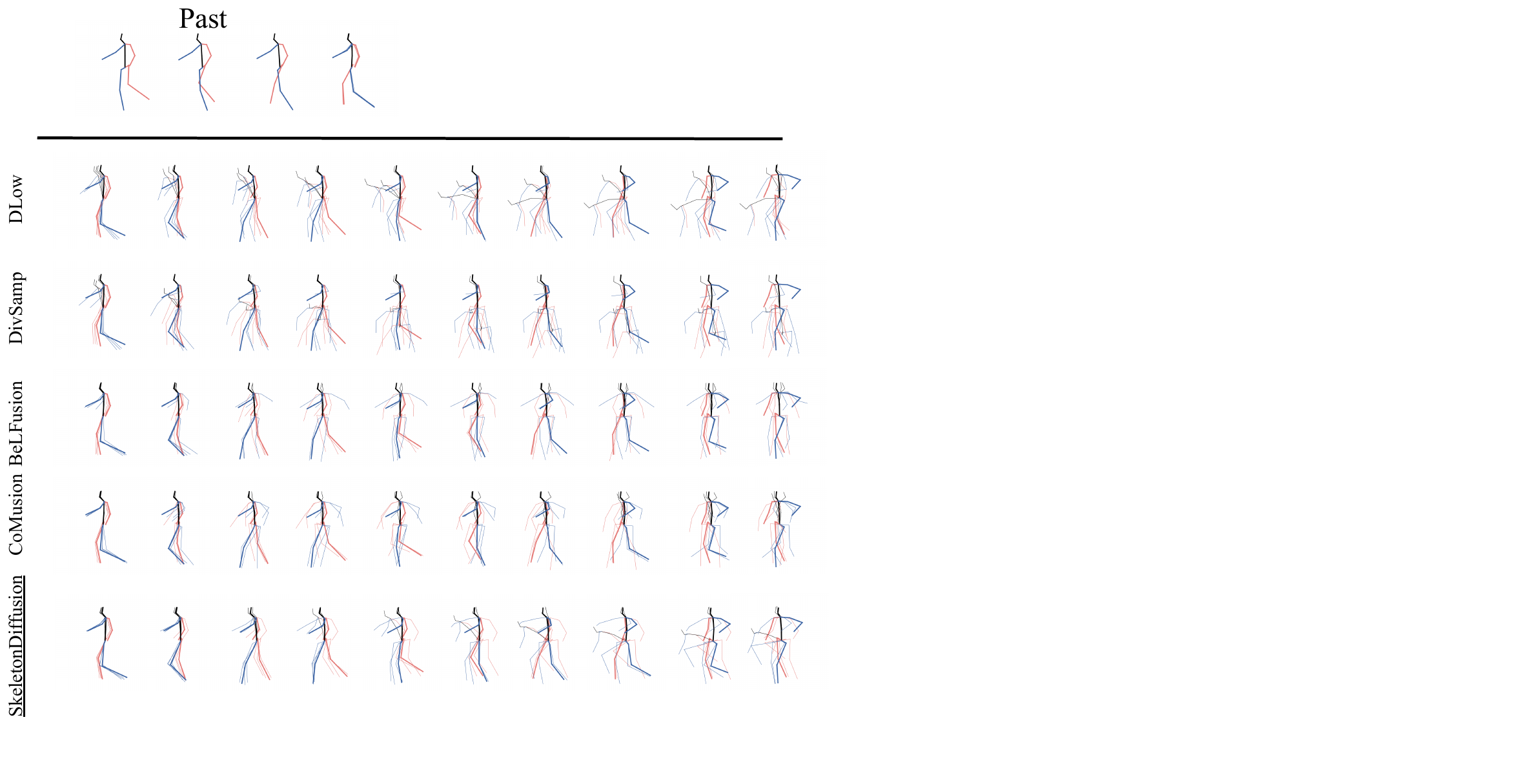}
  \vspace{-0.6cm}
  \caption{Qualitative Results on H36M through overlapping skeletons. Action labeled WalkTogether, segment n. 791. For each method, we display the ground truth future (thicker skeleton) overlapped by the closest prediction and the two most diverse. See \cref{fig:qualitative_h36m3} for a different visualization of the same qualitative.} 
  \label{fig:qualitative_skeleton}
\end{figure}
\section{Training Details}
The conditional autoencoder is trained for 300 epochs on AMASS, 200 on FreeMan, and 100 on H36M.  
In the autoencoder training, to avoid collapse towards the motion mean of the training data \cite{bhattacharyya2018accurate, wang2021simple}, we employ curricular learning \cite{bengio2009curriculum, wang2021simple, adeli2021tripod} and learn to reconstruct sequences with random length $l$, sampled from a discrete uniform distribution $l \sim  {\mathcal {U}}\{1, \tilde{\futurewindow}\}$. 
Specifically, we increase the upper bound of the motion length $\tilde{\futurewindow}$ to the original future timewindow $\futurewindow$ after the first 10 epochs with a cosine scheduler.
The denoiser network is trained with $T=10$ diffusion steps and a learning rate of 0.005 for 150 epochs.
We employ a cosine scheduler \cite{nichol2021improved} for $\alphaT{}{t}$ and implement an exponential moving average of the trained diffusion model with a decay of 0.98.
Inference sampling is drawn from a DDPM sampler \cite{ho2020denoising}. Both networks are trained with Adam on PyTorch. 
The biggest version of our model (AMASS) consists of 34M parameters and is trained on a single NVIDIA GPU A40 for 6 days. For AMASS, we measure an inference time of 471 milliseconds for a single batch on a NVIDIA GPU A40, in line with the latest DM works.

\begin{figure*}[t] \footnotesize	
  \centering
  \includegraphics[width=0.86\textwidth]{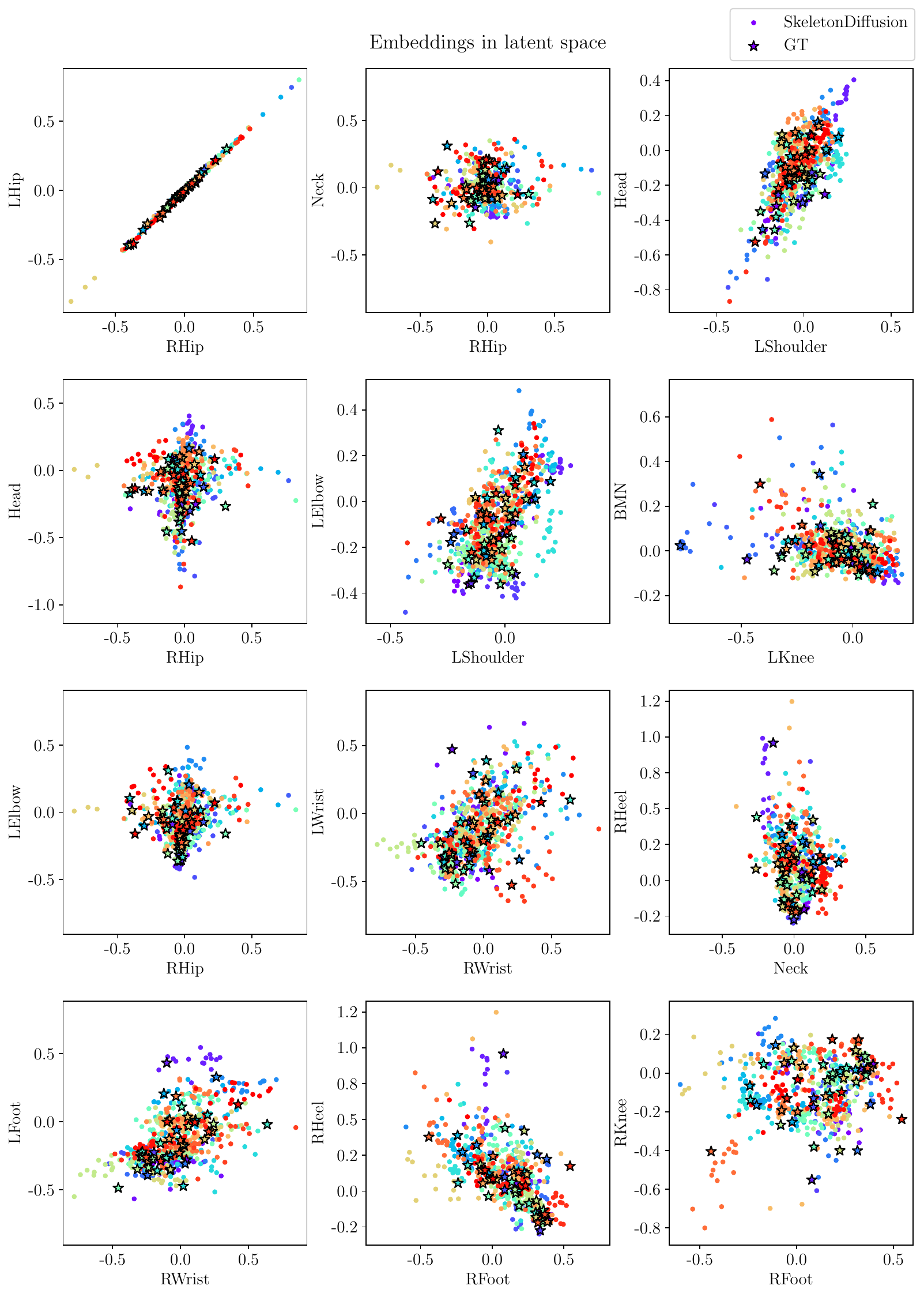}

  \vspace{-0.35cm}
  \caption{PCA plots of latent space embeddings for AMASS GT test segments with corresponding diffused latents generated by \ourmethod. Each GT embedding is denoted by a $\star$ of a different color, and the generated latents corresponding to the same past are denoted by a circle $\circ$ of the same color. %
  }
  \label{fig:LatentSpace_GTembeds}
\end{figure*}

\begin{figure*}
  \centering
  \includegraphics[width=\textwidth]{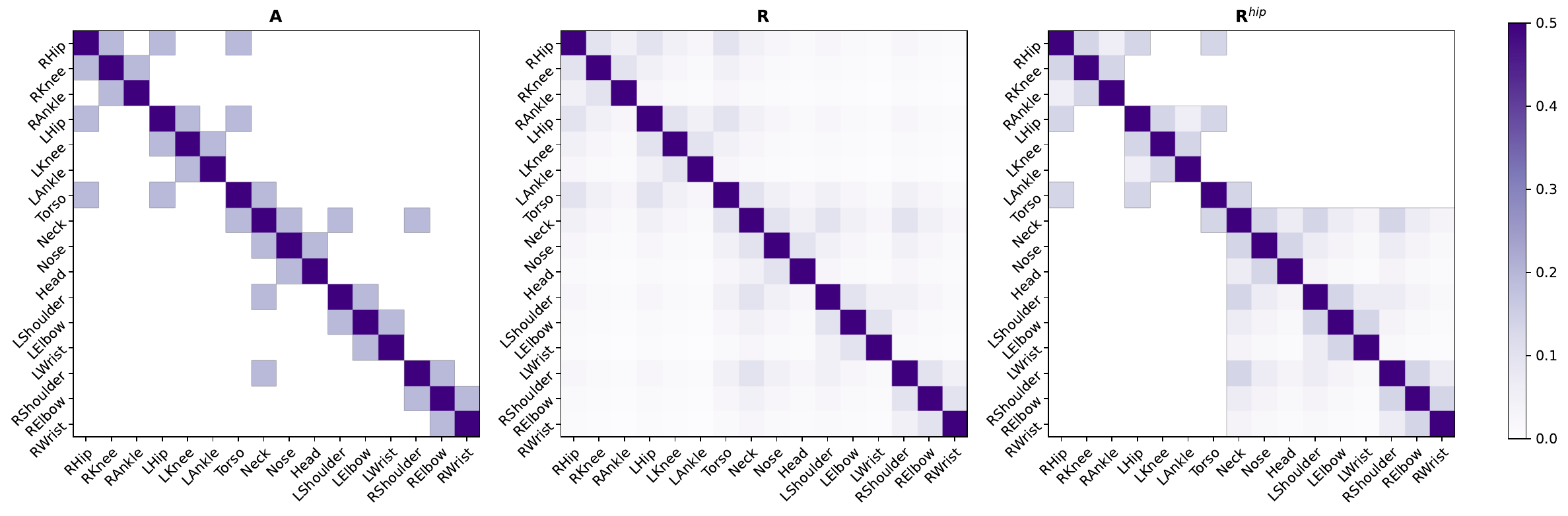}
\vspace{-0.7cm}
  \caption{Node correlation matrix $\covadj$ for different starting choices on the H36M skeleton: the adjacency matrix $\adjmatrix$ of the skeleton graph, the weighted transitive closure $\reachmatrix$ and the masked weighted transitive closure $\reachmatrixhip$.}
  \label{fig:ablation_sigmaN}
\end{figure*}

\begin{table*}\footnotesize	
\centering

\begin{tabular}{l cccccc cccc}
    \toprule
          & \multicolumn{2}{c}{Precision} & \multicolumn{2}{c}{Multimodal GT} & \multicolumn{1}{c}{Diversity} & \multicolumn{1}{c}{Realism} & \multicolumn{4}{c}{{Body Realism}}\\
  \cmidrule(lr){2-3} \cmidrule(lr){4-5} \cmidrule(lr){6-6} \cmidrule(lr){7-7} \cmidrule(lr){8-11}
 \multirow{2}{*}{Base of $\covadj$}  & &  & & & &  & \multicolumn{2}{c}{mean $\downarrow$} & \multicolumn{2}{c}{RMSE $\downarrow$} \\  
 & ADE $\downarrow$ & FDE $\downarrow$ & MMADE $\downarrow$ & MMFDE $\downarrow$ & APD $\uparrow$ & CMD $\downarrow$ &  str  & jit & str  & jit\\
    \midrule
    $\reachmatrix$ & 0.481 & \textbf{0.540} &  0.562 &  \textbf{0.574} & \bfseries{9.504}&  11.542 & 3.16 & 0.20 & 4.51 & 0.27\\ %
    $\reachmatrixhip$ & \bfseries 0.475 & 0.543 & \bfseries 0.558 & 0.579 & 8.629 & 12.499 & \bfseries 3.14 & \bfseries 0.19 & \bfseries 4.35 & \bfseries 0.25 \\
    $\adjmatrix$ (\ourmethod) &  0.480 &  0.545 &  {0.561} & {0.580} & 9.456 & \bfseries{11.417} &  3.15 &  0.20 & 4.45 & 0.26 \\
    \bottomrule
\end{tabular}
    \caption{Ablation studies for the correlation matrix $\covadj$ on AMASS for adjacency matrix $\adjmatrix$, the weighted transitive closure $\reachmatrix$, and the masked weighted transitive closure $\reachmatrixhip$.}
    \label{tab:sigmaN}
\end{table*}

\section{Details on Experiment Settings}

\subsection{Metrics in Stochastic HMP}
\label{app:metrics}
 First, we want to evaluate whether the generated predictions $\tilde{\futurevar{}} \in \mathbb{R}^{\nsamples \times \futurewindow \times \numjoints \times 3}$ include the data ground truth and define \emph{precision} metrics: the Average Distance Error (ADE) measures the Euclidean distance between the ground truth $\futurevar{}$ and the closest predicted sequence
 \begin{equation}
         \mathrm{ADE}(\tilde{\futurevar{}}, \futurevar{}) = \min_{n} \| \tilde{\futurevar{}}^n  - \futurevar{} \|_2 ,
 \end{equation}
 while the Final Distance Error (FDE) considers only the final prediction timestep $F$
\begin{equation}
     \mathrm{FDE}(\tilde{\futurevar{}}, \futurevar{}) = \min_{n} \| \tilde{\futurevar{}}^n_{\futurewindow} - \futurevar{\futurewindow} \|_2 .
 \end{equation}
Because of the probabilistic nature of the task, we want to relate the predicted motions not only to a single (deterministic) ground truth but to the whole ground truth data distribution. To this end, we construct an artificial \textit{multimodal} ground truth (MMGT) \cite{barquero2023belfusion, yuan2020dlow}, an ensemble of motions consisting of test data motions that share a similar last observation frame. For a sample $j$ in the dataset defined by a past observation $\pastvar{}$ and a ground truth future $\futurevar{}^{j}$, if the distance between the last observation frame and the last observation frame of another sample $m$ is below a threshold $\delta$, the future of that sample $m$ is part of the multimodal GT for $j$: 
 \begin{equation}
 {}^{\mathrm{MM}}\futurevar{}^{j} = \{\futurevar{}^m |\, m: \| \pastvar{0}^m - \pastvar{0}^j \|_2 < \delta,\, m \neq j \} 
 \end{equation}
 The \emph{multimodal} versions of the precision metrics (MMADE and MMFDE) do not consider the predicted sequence closest to the ground truth, but the one closest to the MMGT
\begin{align}
    \mathrm{MMADE}(\tilde{\futurevar{}}, {}^{\mathrm{MM}}\futurevar{}) = &  \min_{(i,j) \in \mathcal{M}} \| \tilde{\futurevar{}}^i - {}^{\mathrm{MM}}\futurevar{}^j \|_2 \\
    \mathrm{MMFDE}(\tilde{\futurevar{}}, {}^{\mathrm{MM}}\futurevar{})  = &  \min_{(i,j) \in \mathcal{M}} \| \tilde{\futurevar{F}}^i - {}^{\mathrm{MM}}\futurevar{F}^j \|_2
 \end{align}
  \begin{equation}
     \text{with} \,\, \mathcal{M} =   \{ (i,j) |\,i\in [1\ldots N],\, j \in [1\ldots M]\}.
  \end{equation}
 While evaluation metrics involving the MMGT may have been meaningful in the early stages of SHMP, these values should be contextualized now that methods have achieved a different level of performance: by definition, the MMGT may contain semantically inconsistent matches between past and future, which is a highly undesirable characteristic for a target distribution.
 
Regardless of their similarity with the ground truth data, the generated predictions should also exhibit a wide range of diverse motions. \emph{Diversity} is measured by the Euclidean distance between motions generated from the same observation as the Average Pairwise Distance (APD):
  \begin{equation}
          \mathrm{APD}(\tilde{\futurevar{}}) =  \frac{1}{|\mathcal{P}|} \sum_{(i, j) \in \mathcal{P}} \| \tilde{\futurevar{}}^i - \tilde{\futurevar{}}^j \|_2 \\
    \end{equation}
  \begin{equation}
    \text{with} \,\, \mathcal{P} =  \{ (i,j) |\,i\in [1\ldots N],\, j \in [1\ldots N],\, i \neq j \}.
  \end{equation}
 Diversity can also be seen in relation to the MMGT: the Average Pairwise Distance Error (APDE) \cite{barquero2023belfusion} measures the absolute error between the APD of the predictions and the APD of the MMGT
\begin{equation}
 \mathrm{APDE}(\tilde{\futurevar{}}, {}^{\mathrm{MM}}\futurevar{}) = | \mathrm{APD}(\tilde{\futurevar{}})  - \mathrm{APD}({}^{\mathrm{MM}}\futurevar{})|.
\end{equation}

Generated motions should not only be close to the GT and diverse, but also \emph{realistic}. Barquero et al. \cite{barquero2023belfusion} address realism  in the attempt to identify speed irregularities between consecutive frames:  the Cumulative Motion Distribution (CMD) measures the difference between the average joint velocity of the test data distribution $\bar{M}$ and the per-frame average velocity of the predictions $M_\timesteps$.
 \begin{equation}
    \begin{split}
        \mathrm{CMD} = & \sum_{i=\timesteps}^{F-1} \sum_{f=1}^{\timesteps} \| M_\timesteps - \bar{M} \|_1 \\
         = & \sum_{f=1}^{F-1} (F - f) \| M_\timesteps - \bar{M} \|_1 
    \end{split}     
 \end{equation}
The Fréchet inception distance (FID) is computed for H36M only (as in [4, 13, 62]), as obtaining the necessary classifier to compute the features is not trivial: AMASS does not have class labels (recently, BABEL \cite{punnakkal2021babel} annotated only 1\% of the test data), and FreeMan's annotations do not map into specific classes. 

\subsection{Baselines}
For the comparison on AMASS, H36M, and 3DPW we employ model checkpoints provided by the official code repositories \cite{barquero2023belfusion, chen2023humanmac, suncomusion, wei2023human} or subsequent adaptations \cite{barquero2023belfusion} of older models \cite{walker2017pose, yuan2020dlow, mao2021generating, dang2022diverse}. HumanMac official repository does not provide a checkpoint for AMASS, and hence it has been discarded.  %
For APD on H36M, MotionDiff released implementation uses a different definition which leads to significantly different results. In \cref{tab:main_h36m}, we report the results of their checkpoint evaluated with the same metric we used for other methods.
\begin{table*}\footnotesize	
\centering
\begin{tabular}{l cccccc cccc}
    \toprule
      & \multicolumn{2}{c}{Precision} & \multicolumn{2}{c}{Multimodal GT} & \multicolumn{1}{c}{Diversity} & \multicolumn{1}{c}{Realism} & \multicolumn{4}{c}{{Body Realism}}\\
  \cmidrule(lr){2-3} \cmidrule(lr){4-5} \cmidrule(lr){6-6} \cmidrule(lr){7-7} \cmidrule(lr){8-11}
 \multirow{2}{*}{Norm Type}  & &  & & & &  & \multicolumn{2}{c}{mean $\downarrow$} & \multicolumn{2}{c}{RMSE $\downarrow$} \\  
 & ADE $\downarrow$ & FDE $\downarrow$ & MMADE $\downarrow$ & MMFDE $\downarrow$ & APD $\uparrow$ & CMD $\downarrow$ &  str  & jit & str  & jit\\
    \midrule
    Frob & 0.480 & \textbf{0.539}  &  0.561 &  \textbf{0.575} & \textbf{9.468} &  12.066 & 3.26 &  0.20 & 4.54 & 0.26 \\ %
    Spect (\ourmethod)  &  0.480 &  0.545 & {0.561} & {0.580} & 9.456 & \bfseries{11.417} & \bfseries 3.15 & \bfseries 0.20 & \bfseries4.45 & \bfseries0.26\\
    \bottomrule
\end{tabular}
    \caption{Ablation on the magnitude normalization procedure for $\covadj$ on AMASS. While normalizing with the Frobenius norm and the Spectral norm deliver very similar results, in favor of realism we opt for the spectral norm.  }
    \label{tab:sigmaN_normtype}
\end{table*}

\begin{table*}\footnotesize	
\centering
\begin{tabular}{l c cc cc cc}
    \toprule
       & & \multicolumn{2}{c}{Precision} & \multicolumn{2}{c}{Multimodal GT} & \multicolumn{1}{c}{Diversity} & \multicolumn{1}{c}{Realism} \\
  \cmidrule(lr){3-4} \cmidrule(lr){5-6} \cmidrule(lr){7-7} \cmidrule(lr){8-8} 
 Type & param\# & ADE $\downarrow$ & FDE $\downarrow$ & MMADE $\downarrow$ & MMFDE $\downarrow$ & APD $\uparrow$ & CMD $\downarrow$\\
    \midrule
isotropic&  \multirow{ 2}{*}{9M}  & 0.509 & 0.571& 0.576 & 0.598 & 7.875 & 16.229 \\
\ourmethod &  & 0.493 & 0.554  & 0.565 & 0.585 & 7.865 & 15.767 \\
\midrule
isotropic & \multirow{ 2}{*}{34M} &  0.499 & 0.553 &  0.568 &  0.583 & 8.788 & 15.603\\

\ourmethod &  &  0.480 &  0.545 & {0.561} & {0.580} & 9.456 & 11.417\\ 
\bottomrule
\end{tabular}
        \caption{Effect of parameters number on AMASS for different types of Gaussian diffusion. Our nonisotropic diffusion training requires fewer training parameters than the isotropic formulation to reach comparable performance.}
    \label{tab:diff_num_layers}
    \vspace{-0.3cm}
\end{table*}
 
\begin{figure}
  \centering
  \begin{subfigure}{\linewidth}
     \includegraphics[trim={0cm, 0.2cm, 0cm, 0.2cm},clip,width=\linewidth]{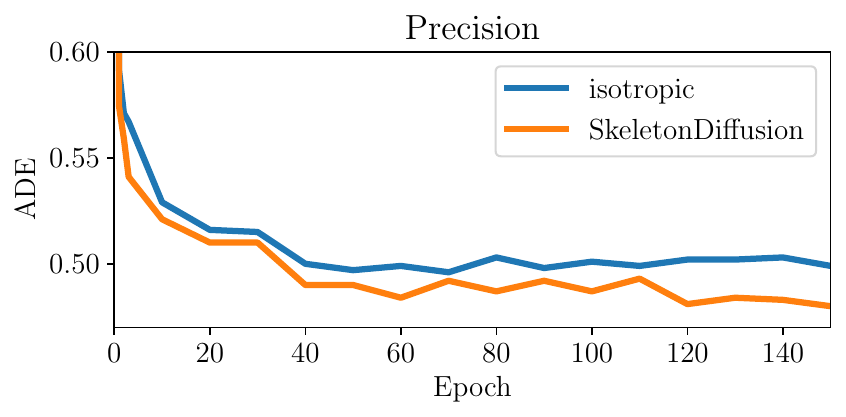}
  \end{subfigure}
    
\begin{subfigure}{\linewidth}
         \includegraphics[trim={0cm, 0.2cm, 0cm, 0.27cm},clip,width=\linewidth]{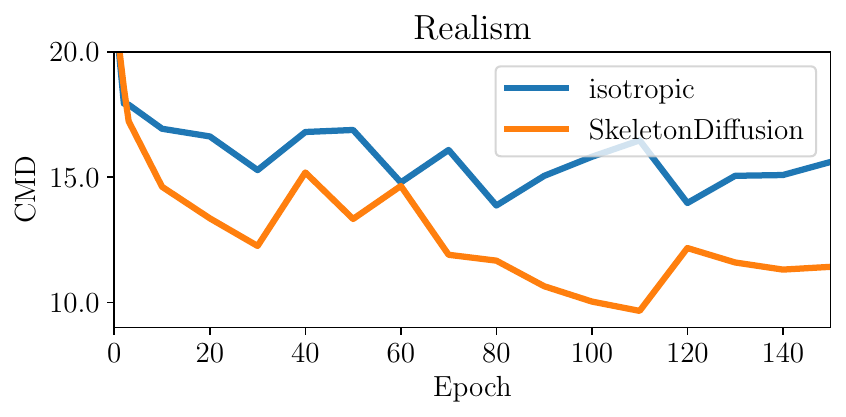}
  \end{subfigure}
  \vspace{-0.8cm}
  \caption{Our nonisotropic diffusion converges in fewer epochs than the conventional isotropic formulation.}
  \label{fig:diff_convergence}
\end{figure}

\begin{table*}[t]\footnotesize	
\centering
\begin{tabular}{l  c ccc cc cccc}
\toprule 
      & \multicolumn{2}{c}{Precision} & \multicolumn{2}{c}{Multimodal GT} & \multicolumn{1}{c}{Diversity} & \multicolumn{1}{c}{Realism} & \multicolumn{4}{c}{{Body Realism}}\\
        \cmidrule(lr){2-3} \cmidrule(lr){4-5} \cmidrule(lr){6-6} \cmidrule(lr){7-7} \cmidrule(lr){8-11}
      & &  & & & &  & \multicolumn{2}{c}{mean $\downarrow$} & \multicolumn{2}{c}{RMSE $\downarrow$} \\  
 & ADE $\downarrow$ & FDE $\downarrow$ & MMADE $\downarrow$ & MMFDE $\downarrow$ & APD $\uparrow$ & CMD $\downarrow$ &  str  & jit & str  & jit\\
\midrule
w/o-TG-Att & 0.502 & 0.567 & 0.576 & 0.597 & 8.021 & 14.934 & 3.90 & 0.20 & 5.31 & 0.27 \\
iso & 0.499 & 0.553 & 0.568 & 0.583 & 8.788 & 15.603 & 3.72 & \bfseries{0.18} & 4.93 & \bfseries{0.24} \\
noniso & \underline{0.489} & \underline{0.547} & \underline{0.567} & \underline{0.581} & 9.483 & \underline{11.812} & \bfseries{2.77} & 0.20 & \bfseries{4.06} & 0.27 \\
Ours (\ourmethod) & \bfseries{0.480} & \bfseries{0.545} & \bfseries{0.562} & \bfseries{0.579} & 9.456 & \bfseries{11.418} & \underline{3.15} & \underline{0.20} & \underline{4.45} & \underline{0.26} \\
\bottomrule
\end{tabular}
\vspace{-0.15cm}
\caption{ Ablations on the AMASS dataset \cite{mahmood2019amass}.} 
\label{tab:ablations}
\vspace{-0.15cm}
\end{table*}
\begin{table*}[t]\footnotesize	
\centering
\begin{tabular}{l  c ccc cc cccc}
\toprule 
      & \multicolumn{2}{c}{Precision} & \multicolumn{2}{c}{Multimodal GT} & \multicolumn{1}{c}{Diversity} & \multicolumn{1}{c}{Realism} & \multicolumn{4}{c}{{Body Realism}}\\
        \cmidrule(lr){2-3} \cmidrule(lr){4-5} \cmidrule(lr){6-6} \cmidrule(lr){7-7} \cmidrule(lr){8-11}
      & &  & & & &  & \multicolumn{2}{c}{mean $\downarrow$} & \multicolumn{2}{c}{RMSE $\downarrow$} \\  
 & ADE $\downarrow$ & FDE $\downarrow$ & MMADE $\downarrow$ & MMFDE $\downarrow$ & APD $\uparrow$ & CMD $\downarrow$ &  str  & jit & str  & jit\\
\midrule
Ours+Past & 0.574 & 0.584 & 0.607 & 0.599 & \underline{9.856} & 16.993 & 10.16 & 0.24 & 11.04 & 0.38 \\
Ours+DCT & 0.534 & 0.572 & 0.595 & 0.600 & \bfseries{11.215} & 16.783 & 5.20 & 0.25 & 7.59 & 0.35 \\
Ours ($k=1$) & 0.489 & 0.601 &  0.577 &   0.639 & 4.984 & 16.574 & 3.50  &  0.17 & 4.56  &  0.23 \\
Ours ($k=50$ latent argmin) & {0.476} & 0.545 &  {0.558} &  0.580 & 8.497 & 12.885 & 3.17  &  0.19 & 4.35 &  0.25 \\
Ours (\ourmethod)& \underline{0.480} & \bfseries{0.545} & \underline{0.562} & \bfseries{0.579} & 9.456 & \bfseries{11.418} & {3.15} & {0.20} & {4.45} & {0.26} \\

\bottomrule
\end{tabular}
\vspace{-0.15cm}
\caption{Additional ablations on AMASS \cite{mahmood2019amass} for discarded components.} 
\label{app:tab:ablations_extra}
\vspace{-0.3cm}
\end{table*}
\begin{figure}
  \centering
   \includegraphics[width=\linewidth]{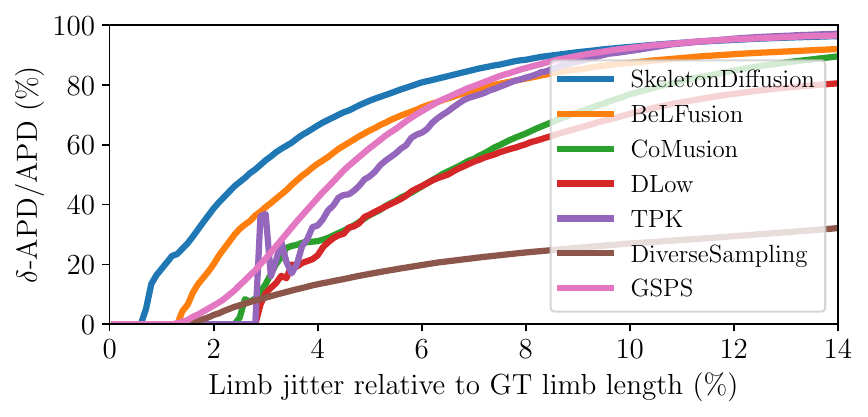}
    \vspace{-0.7cm}
   \caption{Diversity achieved with valid motions over total diversity according to different error tolerances on AMASS.  For every method, we show the evolution of diversity ($
   \delta$-APD) computed with valid motions (y-axis) for which the maximal error is below a given threshold $
   \delta$ (x-axis). \methodname~presents consistently the highest diversity when considering valid poses. } %
   \label{fig:delta-apd}
   \vspace{-0.3cm}
\end{figure}
\subsection{Datasets}
For AMASS, we follow the cross-dataset evaluation protocol proposed by Barquero et al.~\cite{barquero2023belfusion} comprising 24 datasets with a common configuration of 21 joints and a total of 9M frames with 11 datasets for training, 4 for validation, and 7 for testing with 12.7k test segments having a non-overlapping past time window. The MMGT is computed with a threshold of 0.4 resulting in an average of 125 MMGT sequences per test segment.
For 3DPW, we perform zero-shot on the whole dataset merging the original splits, and by employing the same settings as AMASS we obtain 3.2k test segments with an average of 11 MMGT sequences. 
For H36M \cite{Ionescu2014}, as previous works \cite{mao2021generating, salzmann2022motron,dang2022diverse,  barquero2023belfusion, chen2023humanmac, yuan2020dlow, dang2022diverse}, we train with 16 joints on subjects S1, S5, S6, S7, S8 (S8 was originally a validation subject) and test on subjects S9 and S11 with 5.2k segments for an average of 64 MMGT sequences (threshold of 0.5).
FreeMan is a large-scale dataset for human pose estimation collected in-the-wild with a multi-view camera setting, depicting a wide range of actions (such as \textit{pass ball}, \textit{write}, \textit{drink}, \textit{jump rope}, and others) and 40 different actors for a total of 11M frames. 
As FreeMan extracts human poses from RGB, the final data may be noisy and contain ill-posed sequences. We prune the data to obtain fully labeled poses with a limb stretching lower than 5cm, and by applying the same evaluation settings as H36M obtain 11.0k test segments with an average of 69 MMGT. In the next paragraph, we report the pruning protocol.
Note that as FreeMan is collected in the wild, it provides video information that could be potentially used as valuable context information for the human motion prediction task for future works.  
\vspace{-0.5cm}
\paragraph{Pruning Noisy Data on FreeMan}
The authors of FreeMan~\cite{Wang2023freeman} compute 3D keypoints according to different protocols, and we prefer to take the most precise data when available (\textit{smoothnet32} over \textit{smoothnet} over \textit{optim} derivation). The protocols exhibit a restricted number of failure cases (for example, sudden moves very close to camera lenses). 
To avoid training and evaluating on strong failure cases, we remove all sequences where the difference in limb length between consecutive frames in the ground truth exceeds 5cm - a good trade-off between the overall accuracy error range of the dataset and the precision required for the task. 
In comparison, the maximal limb length error between consecutive frames in H36M (MoCap data) is 0.026 mm. Overall we obtain 1M frames, more than three times as much as H36M. 
To balance the splits after pruning, we move test subjects 1, 37, 14, 2, 12 and validation subjects 24, 18, 21 to the train split. We train on 724k densely sampled training segments (3.3k segments for validation). H36M, instead, is composed by 305k samples.

\subsection{Visualization of Generated Motions.} As mentioned in the main paper, often metrics hide or may be influenced by artifacts. Inspecting qualitative results can lead to better insights into the effective SHMP methods' performance. Previous works \cite{barquero2023belfusion, suncomusion, chen2023humanmac, walker2017pose, yuan2020dlow, mao2021generating, dang2022diverse} visualize the diversity of the predictions by overlapping the skeleton of multiple motions in different colors. This representation is limited and not well suited to identify motion irregularities. We propose to fit a SMPL mesh to each skeleton pose to ease inspection of the results, while preserving the semanticity of the prediction. Ill-posed predictions can thus be easily spotted through the erroneous SMPL fitting. For completeness, we still report the historical visualizations in \cref{fig:qualitative_skeleton}.

\section{Further Analysis}

\subsection{Correlations of Latent Space}
\label{appendix:skeleton_latent_space}
We visualize the latent space in terms of the correlation among different latent joint dimensions. To this end, we embed all AMASS test segments in the latent space, and compute the first principal component along the each joint dimension separately. For each embedding, we then plot the principal component of two joint dimensions against each other. In \cref{fig:LatentSpace_GTembeds},  we show 50 random test segments and for each 15 diffused latents. Our latent space reflects correlations connected body joints that are expected (e.g. LHip and RHip) or are less intuitive (e.g. Neck and Hip always show in the same space direction), while other joints do not exhibit univocal correlations (e.g. Wrist and Ankle of the same body side). Weak correlations (probably related to the walking pattern) can be observed between opposite joints of the lower and upper body such as RHip and LElbow.

\begin{table*}[t]\footnotesize	
  \centering
  \setlength{\tabcolsep}{4.0pt}
\begin{tabular}{c  l rrr  rr  rr   rrrrr}
\toprule 
\multicolumn{2}{c}{} & \multicolumn{3}{c}{Precision} & \multicolumn{2}{c}{Multimodal GT} & \multicolumn{1}{c}{Diversity} & \multicolumn{2}{c}{Realism} & \multicolumn{4}{c}{Body Realism}\\
 \cmidrule(lr){3-5} \cmidrule(lr){6-7} \cmidrule(lr){8-8} \cmidrule(lr){9-10} \cmidrule(lr){11-14}
\multirow{2}{*}{Type}  & \multirow{2}{*}{Method}
& & & &  & & & &  &\multicolumn{2}{c}{mean $\downarrow$} & \multicolumn{2}{c}{RMSE $\downarrow$} \\
&   
 & ADE $\downarrow$ & FDE $\downarrow$ & MAE $\downarrow$ & MMADE $\downarrow$ & MMFDE $\downarrow$ & APD $\uparrow$  & CMD $\downarrow$  & FID &  str  & jit & str  & jit\\
\midrule 
Alg & Zero-Velocity   & 0.597 & 0.884 & 6.753 & 0.683 
 & 0.909 & 0.000 &  22.812 & 0.606 &  0.00 &  0.00 &  0.00 &  0.00\\
\midrule
\multirow{2}{*}{GAN}
& HP-GAN \cite{barsoum2017hpgan}    & 0.858 & 0.867 & - & 0.847 & 0.858 & 7.214 & - & - & - & - & - & -\\

& DeLiGAN \cite{gurumurthy2017deligan}   & 0.483 & 0.534 & - & 0.520 & 0.545 & 6.509 & - & - & - & - & - & - \\
\midrule
\multirow{5}{*}{VAE}
& TPK \cite{walker2017pose}   & 0.461 & 0.560 & 8.056 &0.522 & 0.569 & 6.723 & 6.326 & 0.538 & 6.69 & 0.24 & 8.37 & 0.31 \\ %
& DLow \cite{yuan2020dlow}    & 0.425 & 0.518 & 6.856 &0.495 & 0.531 & 11.741 & {4.927} & 1.255 & 7.67 & 0.28 & 9.71 & 0.36 \\ %
& GSPS \cite{mao2021generating}   & 0.389 & 0.496 & 7.171 & 0.476 & 0.525  & {14.757} & 10.758 & 2.103 & 4.83 & 0.19 & 6.17 & 0.24 \\ %
& Motron \cite{salzmann2022motron}   & 0.375 & 0.488 & - & 0.509 & 0.539  & 7.168 & 40.796 & 13.743 & - & - & - & -\\ %
& DivSamp \cite{dang2022diverse}   & {0.370} & 0.485 & 6.257 & {0.475} & {0.516}  & \underline{15.310} & 11.692 & 2.083 & 6.16 & 0.23 & 7.85 & 0.29 \\ %
\midrule
\multirow{ 2}{*}{Other}
& STARS \cite{xu2022diverse}   & 0.358 &\underline{0.445} & - &\underline{0.442}& \underline{0.471} & \textbf{15.884} & - & - & - & - & - & -\\
& SLD \cite{xu2024learning}   & \underline{0.348} & \textbf{0.436} & - & \textbf{0.435} & \textbf{0.463} & 8.741 & - & - & - & - & - & -\\
\hline
\multirow{ 4}{*}{DM}
& MotionDiff \cite{wei2023human}    & 0.411 & 0.509 & - & 0.508 & 0.536 & 7.254 & - & - & 8.04 & 0.59 & 10.21 & 0.77\\ %
& HumanMAC \cite{chen2023humanmac}    & 0.369 & 0.480 & 6.167 & 0.509 & 0.545 & 6.301 & - & - & \underline{4.01} & 0.46 & 6.04 & 0.57\\ %
& BeLFusion \cite{barquero2023belfusion}   & 0.372 & {0.474} & 6.107 &  {0.473} & 0.507  & 7.602 & 5.988 & {0.209} & 5.39 & \underline{0.17} & 6.63 & \underline{0.22} \\ %
 & CoMusion \cite{suncomusion}  & {0.350} & {0.458}  &  5.904 & 0.494 & {0.506}  & 7.632 & \textbf{3.202} & \textbf{0.102} &  4.61 & 0.41 & \underline{5.97} & 0.56 \\ %

 \cmidrule{1-14}\morecmidrules\cmidrule{1-14}
 \multirow{ 1}{*}{DM}
 & SkeletonDiff  & \bfseries 0.344 & {0.450} & 5.556 & {0.487} & 0.512 & 7.249 
 & \underline{4.178} & \underline{0.123} &  \textbf{3.90} & \textbf{0.16}& \textbf{4.96} & \textbf{0.21}\\ %

\bottomrule
\end{tabular}

\caption{Comparison on Human3.6M \cite{Ionescu2014}. Bold and underlined results correspond to the best and second-best results, respectively.}
\label{tab:main_h36m}

\end{table*}

\subsection{Discussion on Correlation Matrix $\covadj$}
\label{appendix:ablation_sigman}

\paragraph{On the Magnitude Normalization}
The magnitude of $\covadj$ is constrained as in \cref{eq:sigman}, where, after adding entries along the diagonal, we divide by the highest eigenvalue (spectral norm). In \cref{tab:sigmaN_normtype}, we show results on AMASS for another normalization choice,  the Frobenius norm i.e. the average of the eigenvalues. While both norms deliver very similar results, we opt for the spectral norm as the realism metrics indicate lower limb stretching and joint velocity closer to the GT data (CMD). An educated guess for the subtle difference is that higher noise magnitude (Frobenius norm) eases the generation of more diverse samples (higher diversity) but at the same time loses details of fine-grained joint positions (lower realism and limb stretching).

\paragraph{Sophistications on the Choice of $\covadj$}
For the correlation matrix $\covadj$ from \cref{eq:sigman}, we opt for the most straightforward and simple starting choice, the adjacency matrix $\adjmatrix$. Here we report further studies to two more sophisticated initial choices:  the weighted transitive closure $\reachmatrix$ and the masked weighted transitive closure $\reachmatrixhip$. 
Given two nodes $v_i$ and $v_j$ in the graph, the shortest path is denoted by $P(i, j)$.
The number of hops between $v_i$ and $v_j$ is denoted by $h_{i,j}$.
We then can express the weighted transitive closure $\reachmatrix$ as:
\begin{equation}
    \reachmatrix_{i,j} := \eta^{h_{i,j}-1}
\end{equation}
with some $\eta \in (0, 1)$, representing the reachability of each node weighted by the hops.
As the hip joint is critical in human motion, we also consider a masked version $\reachmatrixhip$:
\begin{equation}
\begin{aligned}
    \reachmatrix^{hip}_{i,j} = \begin{cases} 
                    \reachmatrix_{i,j} & \text{ if } v_{hip} \in P(i,j), v_i \neq v_{hip}, v_j \neq v_{hip} \\
                    0 & \text{  otherwise} \\
    \end{cases}
\end{aligned}
\end{equation}
These three node correlation matrices are visualized on the H36M dataset in Fig.~\ref{fig:ablation_sigmaN}.  While all three alternatives obtain good results on AMASS in \cref{tab:sigmaN}, we opt for the adjacency matrix $\adjmatrix$ as it is not handcrafted and allows our nonisotropic approach to generalize in a straightforward manner to different datasets. We see the analysis of sophisticated choices for $\covadj$ as an exciting future direction. %

\subsection{On the Convergence of Nonisotropic Diffusion}
\label{supp:exp:noniso_vs_iso}

As depicted in \cref{fig:diff_convergence}, our nonisotropic formulation converges faster than the isotropic counterparty. As the time required for a train iteration is equal among both formulations up to a few negligible matrix multiplications, our nonisotropic formulation achieves higher performance in fewer iterations.
In \cref{tab:diff_num_layers}, we show that for similar performance (precision ADE) our nonisotropic formulation requires fewer parameters than conventional isotropic diffusion. We report these findings as they may be relevant for HMP applications or other structured tasks employing diffusion models.

\begin{table*}[ht]\footnotesize
\centering
\begin{tabular}{ c l  rrr  rr  rr   rrrr}
\toprule
 & & \multicolumn{3}{c}{Precision} & \multicolumn{2}{c}{Multimodal GT} & \multicolumn{1}{c}{Diversity} & \multicolumn{1}{c}{Realism} & \multicolumn{4}{c}{Body Realism}\\
 \cmidrule(lr){3-5} \cmidrule(lr){6-7} \cmidrule(lr){9-9} \cmidrule(lr){8-8} \cmidrule(lr){10-13}
\multirow{2}{*}{Type}  & \multirow{2}{*}{Method} &  & &  & & & &  &\multicolumn{2}{c}{mean $\downarrow$} & \multicolumn{2}{c}{RMSE $\downarrow$} \\

 & & ADE $\downarrow$ & FDE $\downarrow$ & MAE $\downarrow$ & MMADE $\downarrow$ & MMFDE $\downarrow$ & APD $\uparrow$  & CMD $\downarrow$ &  str  & jit & str  & jit\\
\midrule
Alg & ZeroVelocity & 0.764 & 1.016 & 10.921 & 0.785 & 1.019 & 0.000 & 40.695 & {4.52} & {0.00} & {4.52} & {0.00} \\
\hline
\multirow{2}{*}{VAE} & DLow & 0.596 & 0.652 & 9.188 & 0.615 & 0.654 & 13.776 & 12.754 & 8.79 & 0.43 & 11.73 & 0.63 \\
& DivSamp & 0.583 & 0.690 & 10.758 & 0.617 & 0.698 & \bfseries{23.878} & 46.594 & 12.38 & 0.82 & 18.11 & 1.07 \\
\midrule
\multirow{3}{*}{DM} & BeLFusion & \bfseries{0.507} & \underline{0.596} & 9.914 & \bfseries{0.543} & \underline{0.606} & 7.750 & 16.812 & 9.07 & \underline{0.23} & 10.65 & \underline{0.31} \\
& CoMusion & 0.550 & 0.600 & \underline{8.773} & 0.588 & 0.611 & \underline{14.400} & \underline{12.282} & \underline{6.21} & 0.66 & \underline{8.60} & 0.87 \\
& Ours & \underline{0.517} & \bfseries{0.587} & \bfseries{7.106} & \underline{0.567} & \bfseries{0.603} & 10.547 & \bfseries{8.188} & \textbf{4.56} & \textbf{0.22} & \textbf{5.95} & \textbf{0.30} \\
\bottomrule
\end{tabular}
\vspace{-0.16cm}
\caption{\small Models trained on AMASS tested on zero-shot on 3DPW with synthetic noise up to 2cm added to 25\% of the input.} 
\label{app:tab:3dpw_noisy}
\vspace{-0.15cm}
\end{table*}
\begin{table*}[ht]\footnotesize
\centering
\begin{tabular}{ l  r r r rrrr }
\toprule 
 & \multicolumn{3}{c}{Precision} & \multicolumn{4}{c}{{Body Realism}}\\
  \cmidrule(lr){2-4} \cmidrule(lr){5-8} 
  & & &  & \multicolumn{2}{c}{mean $\downarrow$} & \multicolumn{2}{c}{RMSE $\downarrow$} \\
   & ADE $\downarrow$ & FDE $\downarrow$ &  MAE $\downarrow$ &  str & jit &  str & jit \\ 
\midrule
DLow & 0.716 & \underline{0.776} & 12.397 & 7.36 & 0.23 & 9.57 & 0.40 \\
DivSamp & 0.728 & 0.879 & 12.373 & 5.01 & 0.23 & 7.49 & 0.32 \\
\midrule
BeLFusion & \bfseries{0.657} & \bfseries{0.756} & 11.175 & {8.89} & \underline{0.18} & {10.69} & \underline{0.27} \\
CoMusion & 0.670 & 0.792 & \underline{10.215} & \underline{4.56} & 0.33 & \underline{6.28} & 0.46 \\
\ourmethod & \underline{0.660} & 0.779 & \bfseries{9.045} & \bfseries{3.67} & \bfseries{0.14} & \bfseries{4.94} & \bfseries{0.24} \\

\bottomrule
\end{tabular}
\vspace{-0.15cm}
\caption{Long term prediction (5s) on AMASS via autoregression of models traimed to predict 2s. MMGT is undefined in this case.} 
\label{app:tab:long_term}
\vspace{-0.25cm}
\end{table*}

\subsection{Ablations of \ourmethod}
\label{app:ablations}
In \cref{tab:ablations}, we report the ablations discussed in \cref{results:ablations}. We compare the effect of TG-Attention layers on isotropic diffusion ($\covadj = I$ and $\gammaT{}{t}=0$) and analyze nonisotropic diffusion with our covariance reflecting joint connections $\covadj$ (\cref{eq:sigman}) in the variant where $\gammaT{}{t}=1$ (as in \cref{eq:purediff}) and our blending with the scheduler $\gammaT{}{t}$ (\cref{eq:diff-covt}). 

In \cref{app:tab:ablations_extra}, we report further experiments, such as fine-tuning the encoder responsible for embedding the past observation (\textit{Ours+Past}) or representing motion data via the Discrete Cosine Transform (DCT) \cite{chen2023humanmac}.
From the low precision results of the DCT experiment (\textit{Ours+DCT}) and referring to \cref{tab:main_amass}, we speculate that while DCT seems suitable for transformer-based diffusion models operating in input space \cite{suncomusion, chen2023humanmac}, extracting features directly from Euclidean motion space seems a better choice for latent diffusion models (BeLFusion \cite{barquero2023belfusion} and our method).

\paragraph{Diffusion Training Objective and \textit{k}-Relaxation}
\label{appendix:ablation_k}
In the same table \cref{app:tab:ablations_extra}, we also ablate the relaxation of the diffusion objective described in \cref{method:skeleton-diffusion} \cref{eq:loss2} with three experiments: (1) by not relaxing the diffusion loss during training i.e. setting $k=1$ and backpropagating the loss through the first and only sample; (2) by sampling $k=50$ times and backpropagating through the generated sample that is most similar to the ground truth in \textit{latent space}; and (3) doing so with $k=50$ but in motion space - which is the choice for \ourmethod. 
By looking at the result for the first two cases, we see that $k=1$ generates considerably less diverse futures, confirming the detailed investigation of BeLFusion~\cite{barquero2023belfusion} on how increasing  $k$  leads to higher diversity.  In our case,  setting $k=50$ does not only lead to a double as high diversity score, but also improves precision and realism. We thus believe relaxation to be a strong guidance towards the target distribution, particularly for latent space models. Intuitively, it allows for better coverage of the different future modes of the real data distribution. In \ourmethod, we go a step further and choose the sample to backpropagate the loss not via similarity in latent space, but in motion space. This design choice improves diversity by more than 10\% (!). Intuitively, in the early stage of the diffusion training, the denoiser generates coarse latent codes, whose similarity to the ground truth embedding may 'erroneously' not reflect the effective perceptual similarity in motion or input space, leading to suboptimal training.  We note, though, that this choice leads to considerably increased training time. Due to the recurrent nature of the decoder, decoding the $k=50$ generated samples in motion space leads to more than doubled training time: the diffusion network of \ourmethod~trains in 5 days, while the version with $k=1$ in a single day, and $k=50$ in latent space in .ca a day and half. 

\section{Additional Experiments}

\subsection{Diversity and Body Realism}
\label{appendix:body_realism}
In the main paper we discuss our intuition on how artifacts in the generated motions may lead to increased distance between the predictions and so to a better diversity metric (APD). We wish to provide evidence of this phenomenon with an argument similar to the one employed in \cref{fig:valid_samp_jitter} of the main paper i.e. by inspecting the evolution of the APD metric at different tolerance thresholds of limb jitter.
First, we compute the valid motions among the generated predictions per method on the AMASS dataset, discarding a sequence if it displays a bone length jitter above a given threshold $\delta$. %
By calculating the average pairwise distance APD only between valid motions and relating this value to the customary APD, in \cref{fig:delta-apd} we can see the contribution of ill-posed motions on diversity. Such evolving diversity  differs significantly from the values reported in \cref{tab:main_amass}. Our method generates by a large margin the most diverse motions when considering realism according to limb jitter, demonstrating excellence also under strict constraints. Non-smooth curve regions display the influence of ill-posed motions on diversity when considering a small ensemble of predictions, as for CoMusion and TPK. When the number of valid motions is small and some of them present stretching, removing the unrealistic motions may considerably improve or worsen the average pairwise distance, resulting in sudden jumps in the curves. We are thus the first to demonstrate quantitatively that unrealistic motions increase diversity.

\begin{table}[t!]\footnotesize	
\vspace{0.4cm}
\centering
\begin{tabular}{lrrr}
\toprule
& Memory$\downarrow$ & NumParams$\downarrow$ & Time$\downarrow$  \\
\midrule
 DLow  & 31 MB & 8.1 M & 111 ms   \\
 DivSamp   & 88 MB & 23.1 M & 8 ms    \\
 BeLFusion  & 53 MB & 17.8 M & 10\,341 ms    \\
 HumanMAC   & 114 MB & 28.7 M & 7\,438 ms  \\
 Comusion  & 87 MB & 19 M & 153 ms
 \\
 \midrule
 \ourmethod  & 106 MB & 26.5 M & 412 ms   \\
\bottomrule
\end{tabular}
\vspace{-0.15cm}
\caption{Footprint for a single H36M inference (RTX 6000)}
\label{app:tab:computational_efficiency}
\vspace{-0.3cm}
\end{table}

\subsection{Human3.6M}
\label{appendix:h36m}
In \cref{tab:main_h36m}, we report quantitative results on H36M. The H36M dataset is particularly small and contains only 7 subjects. We consider this dataset less informative about generalization capabilities of the methods, and more vulnerable to overfitting. With analogous considerations as on AMASS, \ourmethod~ achieves state-of-the-art performance. Thanks to the explicit bias on the human skeleton, \ourmethod~consistently achieves the best body realism, in particular in regard to limb stretching. Even in a setting with limited data, the prior on the skeleton structure contributes to achieving consistent realism.

Overall, the body realism metrics for DM methods appear improved compared to AMASS (\cref{tab:main_amass}). 
 Along VAE and DM approaches, another line of work  relies on representation learning and vocabulary techniques \cite{xu2024learning, xu2022diverse}. While these methods achieve good performance, they employ carefully handcrafted loss functions, limiting the angles and bones between body joints or leveraging the multimodal ground truth in loss computations. Inconveniently, they are required to scrape the whole training data to compute the reference values or the multimodal ground truth, with computational expenses that scale quadratically with the number of instances in the dataset and require considerable engineering effort to be adapted to big data.

\subsection{Challenging Scenario: Synthetic Noise in Zero-Shot Generalization}
\label{app:challenging_scenario}
We perform further experiments on the out-of-distribution, in-the-wild data of 3D Poses in the wild (3DPW), evaluated in \cref{tab:main_3dpw}, by designing a challenging scenario with synthetic noise (\cref{app:tab:3dpw_noisy}). We add random noise of a maximal magnitude of 2cm to 25\% of the input observation keypoint, thus testing robustness to noise for models that were trained with precise, MoCap data (AMASS). While the experiments in \cref{tab:main_freeman} show models trained on noisy data (FreeMan), here we test robustness to noise in a zero-shot setting. \ourmethod delivers among the highest precision and diversity, and the most realistic motions with a gap between 26\% and 65\% compared to the otherwise closest competitor, CoMusion (see \cref{tab:main_amass}). While BeLFusion shows jitter values close to ours, the limb stretching and the CMD are almost double as high, meaning that the length of their limbs highly varies over the whole prediction timespan, and the joint velocities are unrealistic: they achieve high precision with extremely unrealistic motions.

\subsection{Long Term Prediction}
We test models trained on AMASS to predict the next 2s in the generation of 5s motions via autoregression (\cref{app:tab:long_term}). Here we focus on Precision and Realism, as the multimodal GT is ill-defined in this setting, and diversity evaluation loses meaning and its measuremnt is polluted by the difficulty of the task. We achieve again the highest realism and SoTA precision demonstrating the effectivness of our explicit bias on the human skeleton.

\subsection{Computational Efficiency}
Measurement are reported in \cref{app:tab:computational_efficiency}. While there is no obvious computational difference between diffusion models in latent (BeLFusion, Ours) and input space (HumanMAC, CoMusion), latent models achieve much better body realism, particularly jitter (\cref{tab:main_amass}), by not working with 3D coordinates directly.

\section{More Qualitative Examples}
\label{app:qualitative}
\begin{figure*}[t]
  \centering
    \vspace{0.0cm}
    \includegraphics[trim={0cm, 6.5cm, 0cm, 7cm},clip,width=\textwidth]{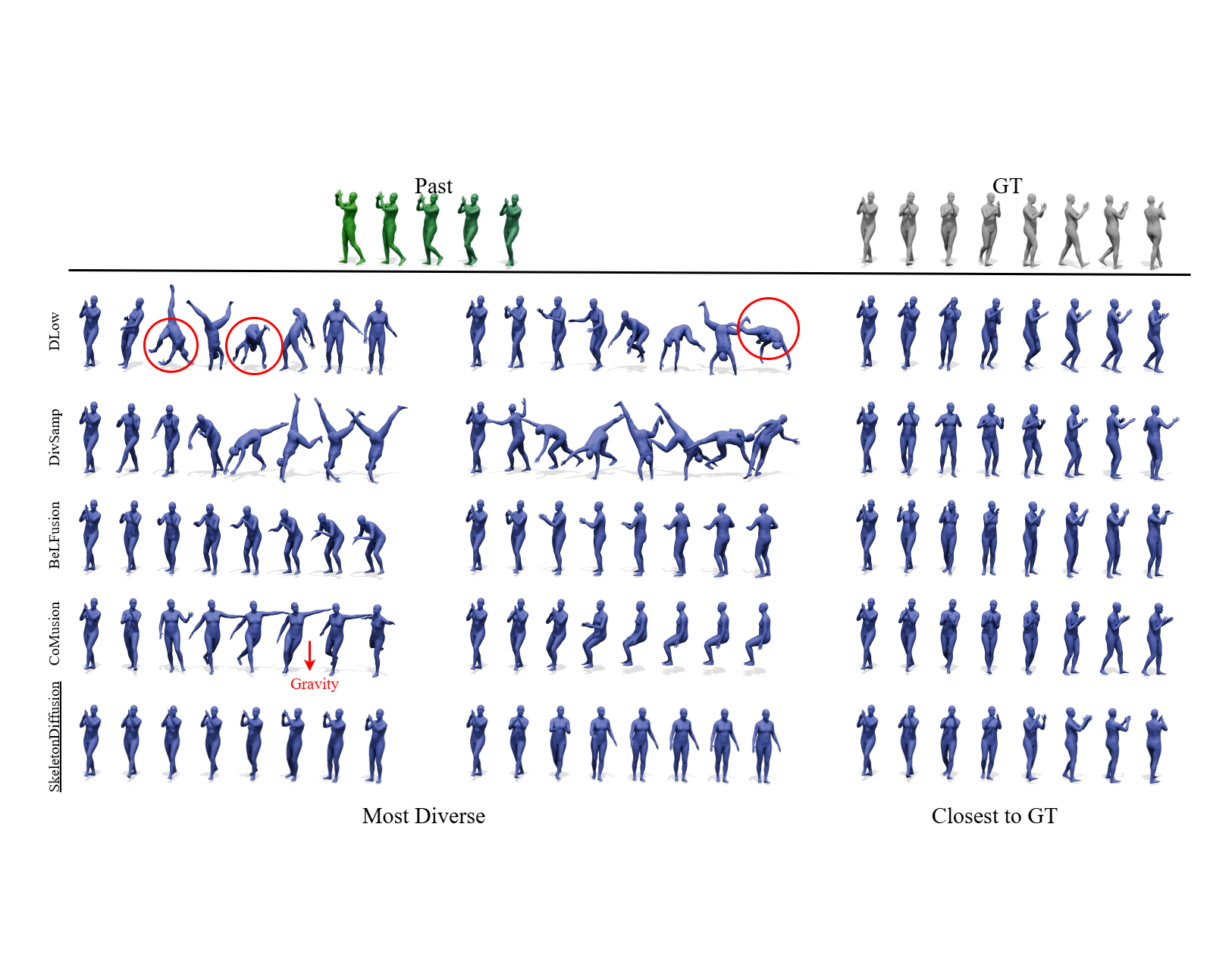}
  \vspace{-0.4cm}
  \caption{Qualitative Results on AMASS. From DanceDB dataset, segment n. 4122.}
  \label{fig:qualitative_amass1}
  \vspace{-0.0cm}
\end{figure*}

\begin{figure*}[t]
  \centering
    \includegraphics[trim={0cm, 6.5cm, 0cm, 7cm},clip,width=\textwidth]{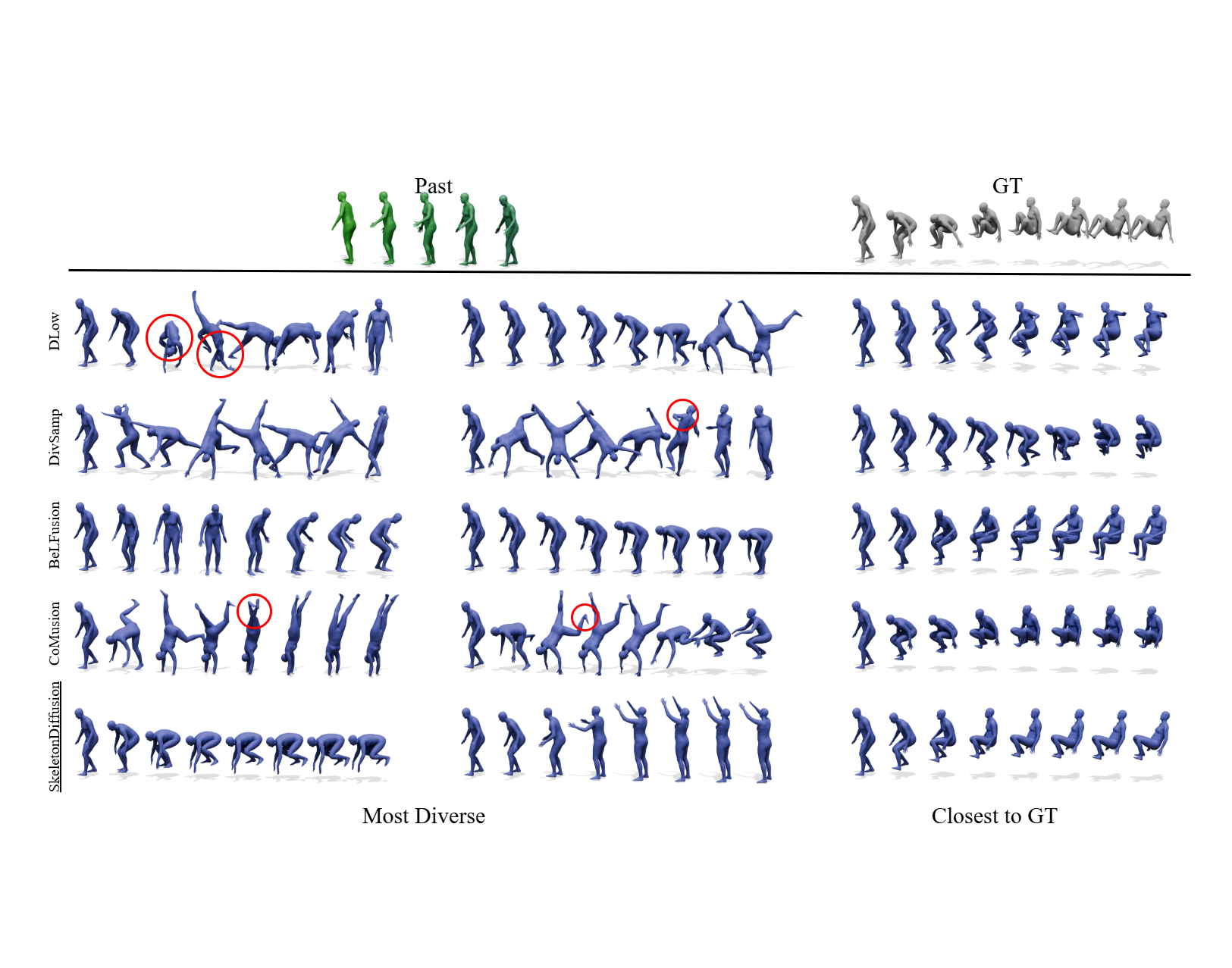}
  \vspace{-0.4cm}
  \caption{Qualitative Results on AMASS. From Human4D dataset, segment n. 11949.}
  \label{fig:qualitative_amass2}
  \vspace{-0.0cm}
\end{figure*}

\begin{figure*}[t]
  \centering
    \vspace{0.0cm}
    \includegraphics[trim={0cm, 6.5cm, 0cm, 7cm},clip,width=\textwidth]{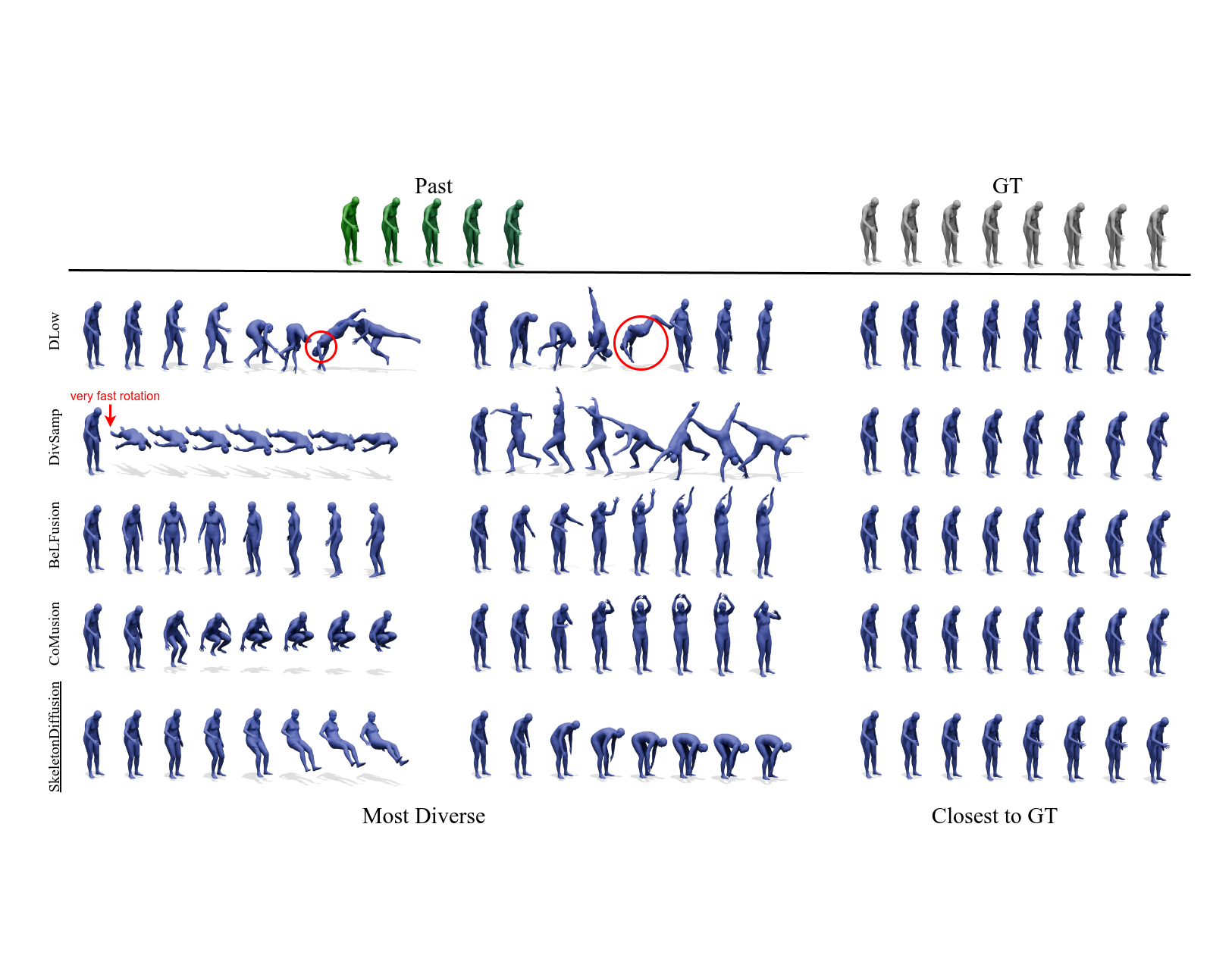}
  \vspace{-0.4cm}
  \caption{Qualitative Results on AMASS. From GRAB dataset, segment n. 9622.}
  \label{fig:qualitative_amass3}
  \vspace{-0.0cm}
\end{figure*}

\begin{figure*}[t]
  \centering
    \includegraphics[trim={0cm, 6.5cm, 0cm, 7cm},clip,width=\textwidth]{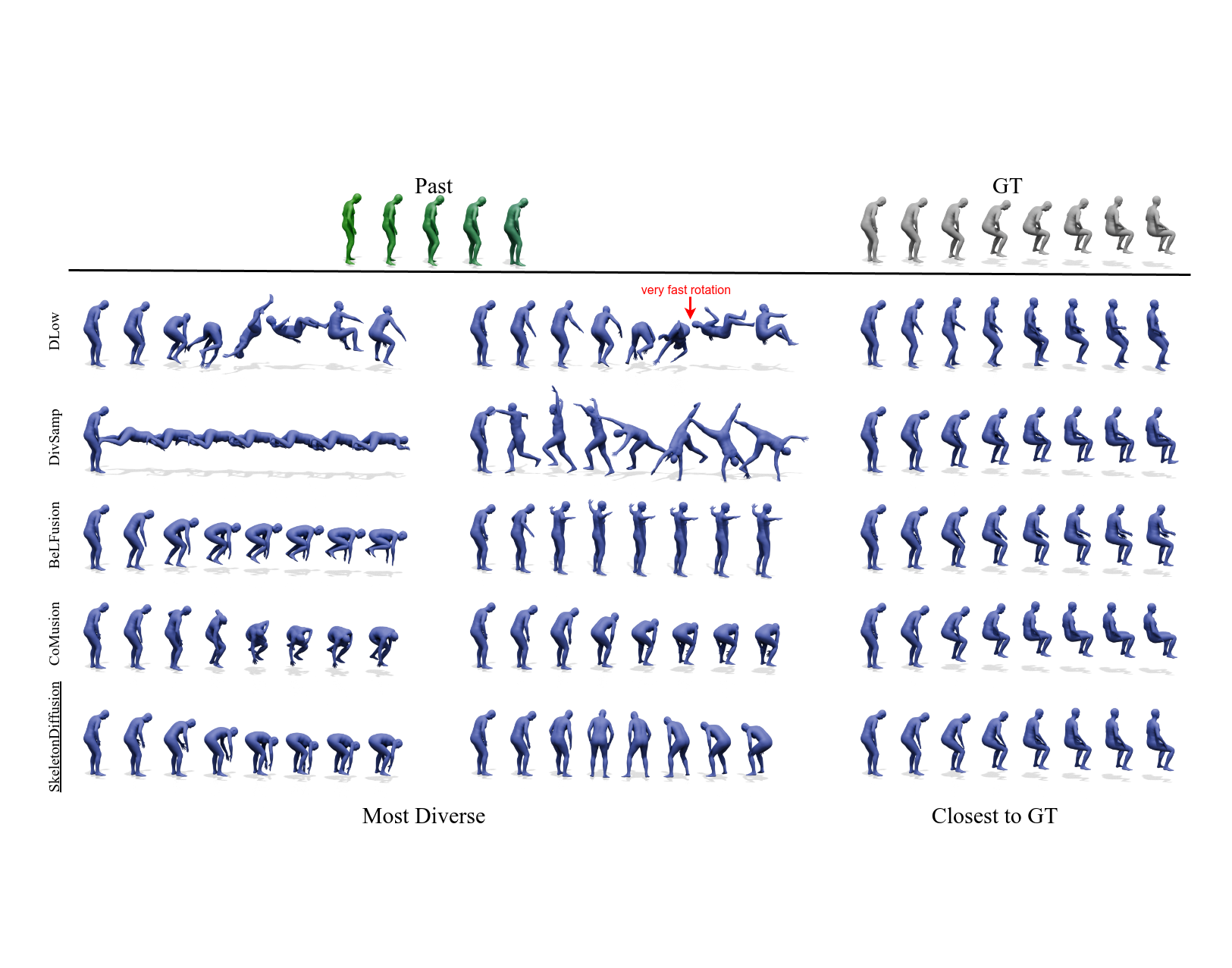}
  \vspace{-0.5cm}
  \caption{Qualitative Results on AMASS. From Human4D dataset, segment n. 12267.}
  \label{fig:qualitative_amass4}
  \vspace{-0.0cm}
\end{figure*}

\begin{figure*}[t]
  \centering
    \vspace{0.0cm}
    \includegraphics[trim={0cm, 6.5cm, 0cm, 7cm},clip,width=\textwidth]{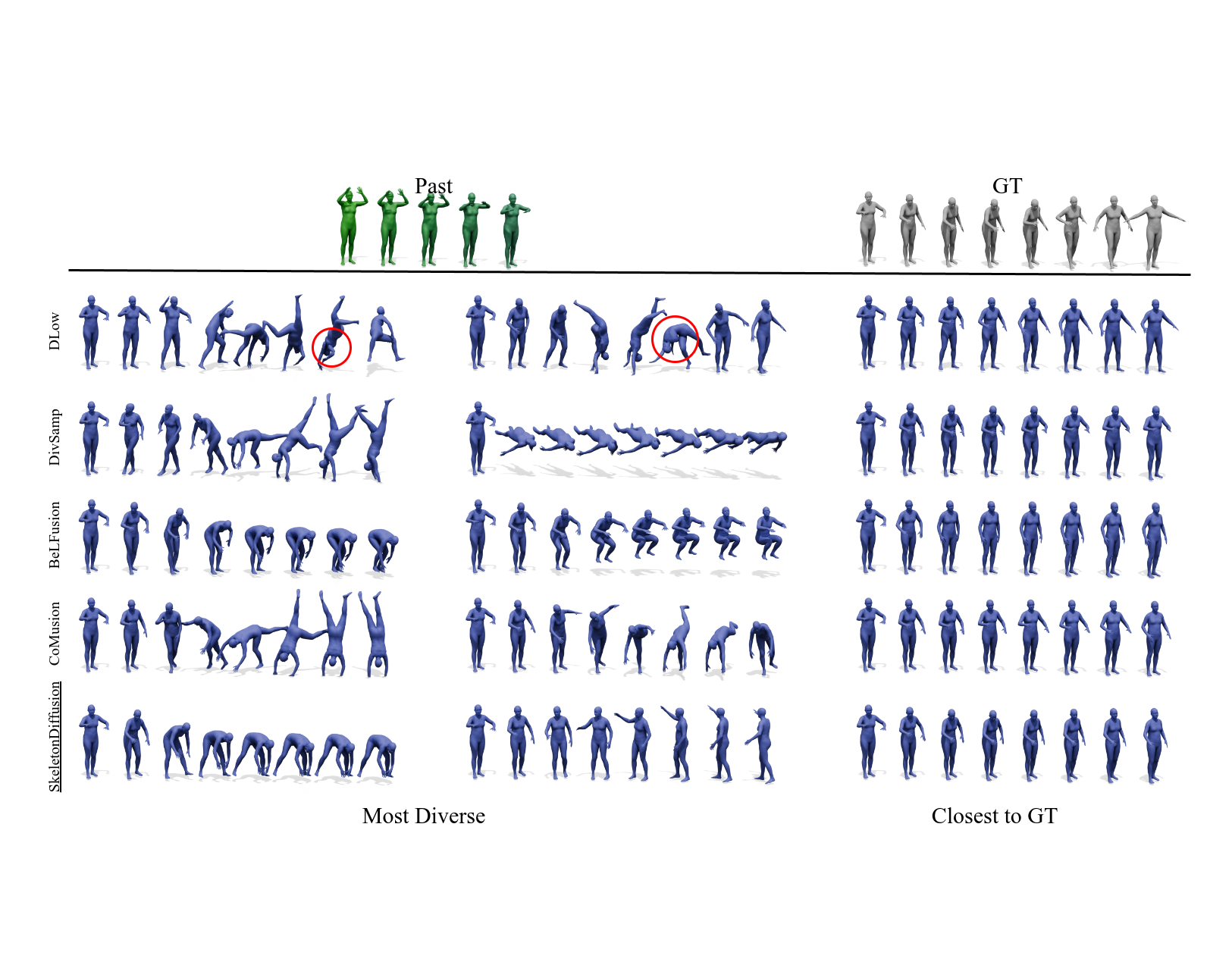}
  \vspace{-0.5cm}
  \caption{Qualitative Results on AMASS. From GRAB dataset, segment n. 10188.}
  \label{fig:qualitative_amass5}
  \vspace{-0.0cm}
\end{figure*}

\begin{figure*}[t]
  \centering
    \vspace{0.0cm}
    \includegraphics[trim={0cm, 6.5cm, 0cm, 8cm},clip,width=\textwidth]{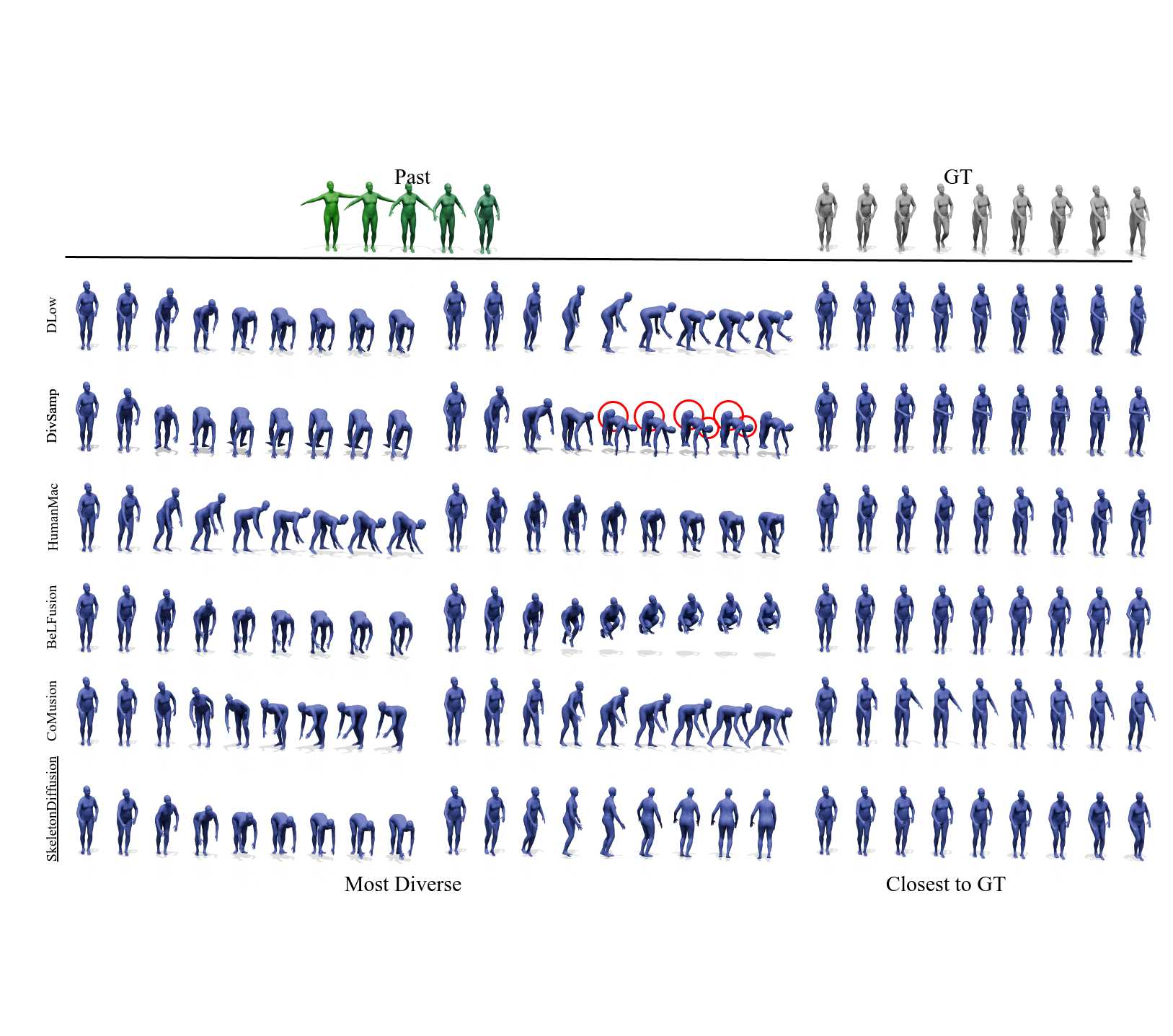}
  \vspace{-0.5cm}
  \caption{Qualitative Results on H36M. Action labeled WalkDog, segment n. 3122.}
  \label{fig:qualitative_h36m1}
  \vspace{-0.0cm}
\end{figure*}

\begin{figure*}[t]
  \centering
    \includegraphics[trim={0cm, 6.5cm, 0cm, 8cm},clip,width=\textwidth]{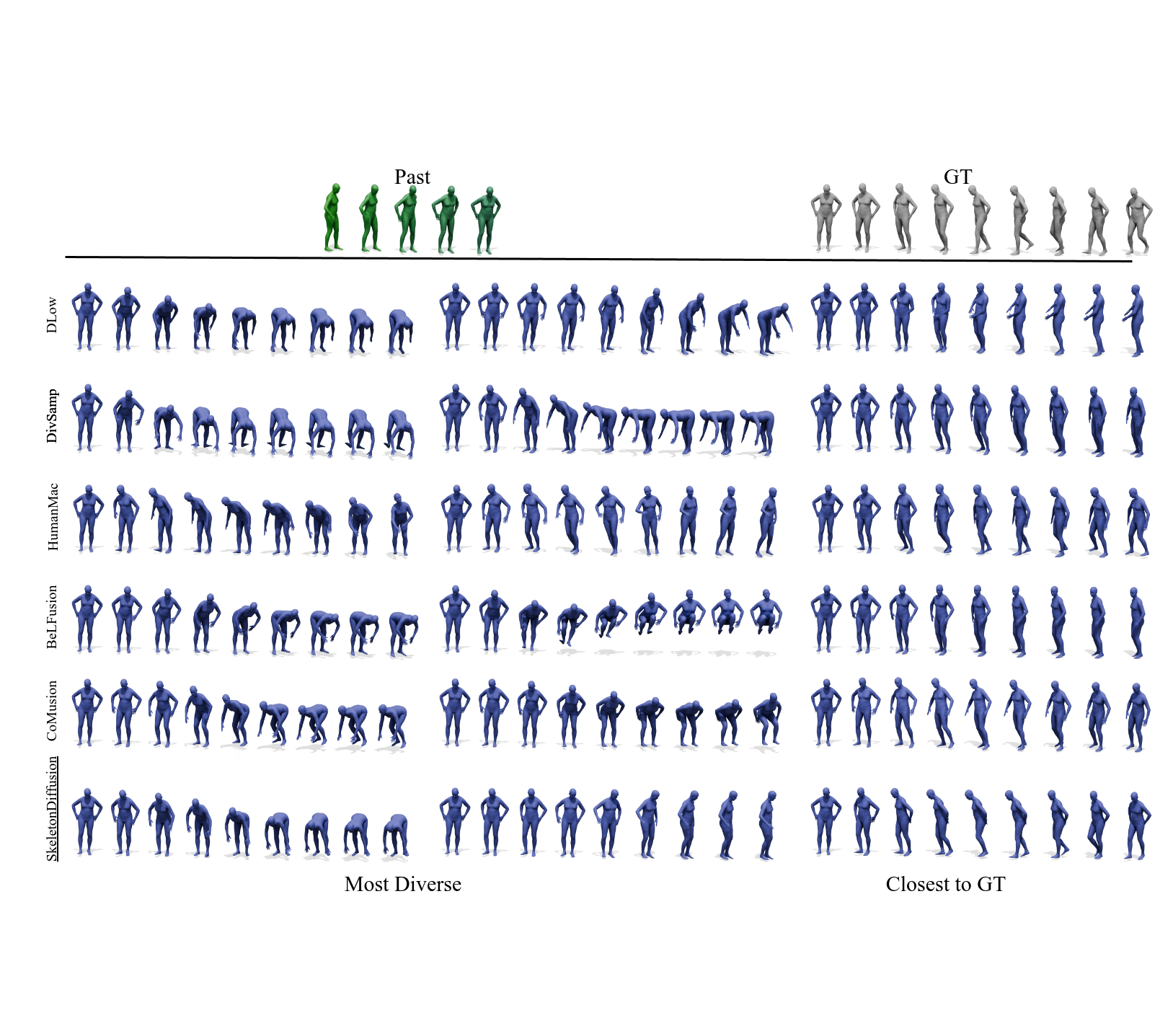}
  \vspace{-0.5cm}
  \caption{Qualitative Results on H36M. Action labeled Discussion, segment n. 2620.}
  \label{fig:qualitative_h36m2}
  \vspace{-0.0cm}
\end{figure*}

\begin{figure*}[t]
  \centering
    \includegraphics[trim={0cm, 6.5cm, 0cm, 8cm},clip,width=\textwidth]{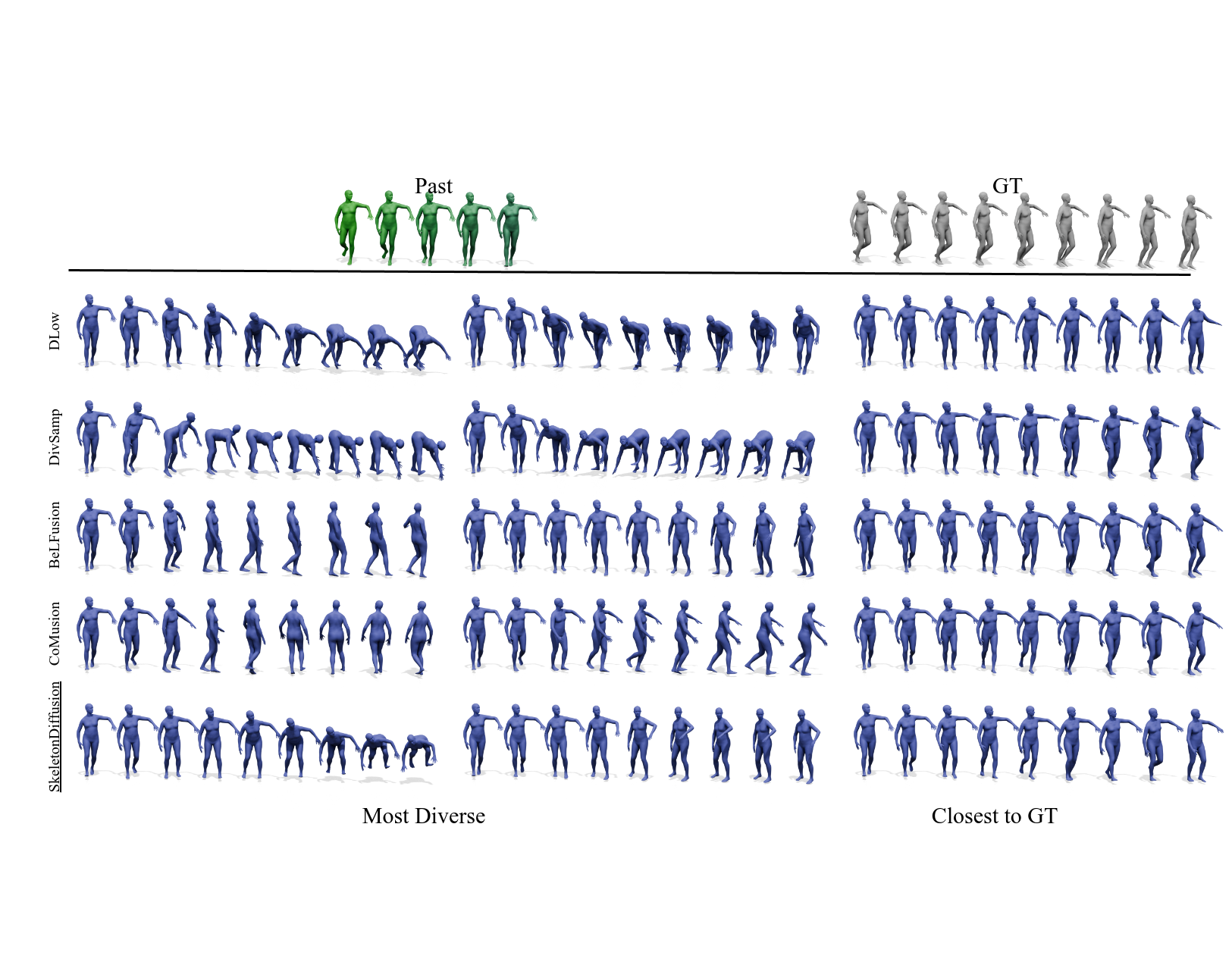}
  \vspace{-0.5cm}
  \caption{Qualitative Results on H36M. Action labeled WalkTogether, segment n. 791.}
  \label{fig:qualitative_h36m3}
  \vspace{-0.0cm}
\end{figure*}

We show more qualitative results on AMASS in \cref{fig:qualitative_amass1,,fig:qualitative_amass2,,fig:qualitative_amass3,,fig:qualitative_amass4,,fig:qualitative_amass5}. More qualitative examples for H36M can be found in  \cref{fig:qualitative_h36m1,,fig:qualitative_h36m2,,fig:qualitative_h36m3} and \cref{fig:qualitative_skeleton}.

\end{document}